\documentclass[lettersize,journal]{IEEEtran}

\usepackage[T1]{fontenc}
\usepackage{float}
\usepackage{xespotcolor}

\usepackage{amsmath,amssymb,amsfonts}
\usepackage{algorithmic}
\usepackage{textcomp}
\usepackage{siunitx}
\usepackage{textcase}

\usepackage[backend=biber,doi=false,isbn=false,eprint=false,style=ieee,url=false]{biblatex}

\usepackage{jabbrv} 

\DeclareFieldFormat[misc]{title}{\mkbibquote{#1}}

\AtEveryBibitem{\clearname{editor}} 
\AtEveryBibitem{\clearfield{series}}
\AtEveryBibitem{\clearfield{issn}}
\AtEveryBibitem{\clearfield{location}} 
\AtEveryBibitem{\clearfield{place}} 
\AtEveryBibitem{\clearlist{language}}
\AtEveryBibitem{\clearfield{pages}}
\AtEveryBibitem{\clearfield{volume}}
\AtEveryBibitem{\clearfield{number}}
  
\AtEveryBibitem{
\ifentrytype{book}{}{
  \clearlist{publisher}
 }
}

\DeclareBibliographyDriver{article}{%
  \usebibmacro{author}%
  \setunit{\addcomma\space}%
  \usebibmacro{title}%
  \setunit{\addcomma\space}%
  \usebibmacro{journal}%
  \setunit{\addcomma\space}%
  \printfield{volume}%
  \setunit{\addcomma\space}%
  \printfield{number}%
  \setunit{\addcomma\space}%
  \printfield{year}%
  \setunit{\addcomma\space}%
  \printfield{pages}%
  \finentry%
}

\DeclareBibliographyDriver{misc}{%
  \usebibmacro{author}%
  \setunit{\addcomma\addspace}%
  \usebibmacro{title}%
  \setunit{\addcomma\addspace}%
  \printfield{year}%
  \setunit{\addcomma\addspace}%
  \printfield{note}%
  \setunit{\addcomma\addspace}%
  \usebibmacro{eprint}%
  \finentry%
}

\bibliography{references_zotero_sahin}

\usepackage{hyperref}
\usepackage{cleveref}

\usepackage{graphicx}
\usepackage[rgb,dvipsnames]{xcolor}
\usepackage{tikz}
\graphicspath{{Figures/hm/}}
\usepackage{svg}
\usepackage{booktabs}
\usepackage{multirow}
\usepackage{multicol}
\usepackage{makecell}    
\usepackage{xargs}
\usepackage{cuted} 
\usepackage{caption}
\usepackage{subcaption}
\usepackage{enumitem}
\usepackage{comment}
\usepackage{fontawesome}

\usepackage[colorinlistoftodos,prependcaption]{todonotes}
\usepackage[normalem]{ulem}
\colorlet{BLACK}{black}

\usepackage[acronym]{glossaries}
\glsdisablehyper
\makeglossaries

\newacronym{ad}{AD}{Autonomous Driving}
\newacronym{genai}{GenAI}{generative AI}

\newacronym{vae}{VAE}{Variational Autoencoder}
\newacronym{gt}{GT}{Generative Transformer}
\newacronym{dm}{DM}{Diffusion Model}
\newacronym{llm}{LLM}{Large Language Model}
\newacronym{ebm}{EBM}{Energy Based Model}
\newacronym{vit}{ViT}{Vision Transformer}
\newacronym{ddpm}{DDPM}{Denoising Diffusion Probabilistic Model}
\newacronym{vlm}{VLM}{Vision Language Model}

\newacronym{node}{NODE}{Neural Ordinary Differential Equation}
\newacronym{ode}{ODE}{Ordinary Differential Equation}
\newacronym{nf}{NF}{Normalizing Flow}
\newacronym{cnf}{CNF}{Continuous Normalizing Flow}
\newacronym{fm}{FM}{Flow Matching}

\newacronym{gan}{GAN}{Generative Adversarial Network}
\newacronym{inn}{INN}{Invertible Neural Network}
\newacronym{cnn}{CNN}{Convolutional Neural Network}
\newacronym{cgan}{CGAN}{Conditional Generative Adversarial Network}
\newacronym{dnn}{DNN}{Deep Neural Network}
\newacronym{rnn}{RNN}{Recurrent Neural Network}
\newacronym{lstm}{LSTM}{Long Short-Term Memory}

\newacronym{sl}{SL}{Supervised Learning}
\newacronym{usl}{USL}{Unsupervised Learning}
\newacronym{rl}{RL}{Reinforcement Learning}
\newacronym{irl}{IRL}{Inverse Reinforcement Learning}
\newacronym{aim}{AIM}{Adversarial Imitation Learning}
\newacronym{il}{IL}{Imitation Learning}
\newacronym{drl}{DRL}{Deep Reinforcement Learning}
\newacronym[
  longplural={Markov Decision Processes}
]{mdp}{MDP}{Markov Decision Process}
\newacronym{bc}{BC}{Behavior Cloning}
\newacronym{e2e}{E2E}{End-to-End}

\newacronym{mc}{MC}{Monte Carlo}
\newacronym{swag}{SWAG}{Stochastic Weight Averaging Gaussian}
\newacronym{hd}{HD}{High Definition}
\newacronym{bev}{BEV}{Bird's-Eye-View}

\newacronym{rag}{RAG}{Retrieval-Augmented-Generation}
\newacronym{knn}{KNN}{K-nearest Neighbors}

\newacronym{nlp}{NLP}{Natural Language Process}
\newacronym{stl}{STL}{Signal Temporal Logic}
\newacronym{mpc}{MPC}{Model Predictive Control}
\newacronym{mppi}{MPPI}{Model Predictive Path Integral}
\newacronym{cbf}{CBF}{Control Barrier Function}
\newacronym{pdm}{PDM}{Predictive Driver Model}
\newacronym{sde}{SDE}{Stochastic Differential Equation}
\newacronym{sotif}{SOTIF}{Safety Of The Intendend Functionality}

\newacronym{nll}{NLL}{Negative Log Likelihood}

\newacronym{sota}{SOTA}{state-of-the-art}

\Crefname{paragraph}{Paragraph}{Paragraphs}

\begin{document}

\title{Generative AI for Autonomous Driving: A Review}

\author{
{Katharina~Winter}$^{1}$, %ORCID ID: 0009-0002-1952-6558
{Abhishek~Vivekanandan}$^{2}$, %ORCID ID: 0009-0007-9834-5873
{Rupert~Polley}$^{2}$, %ORCID ID: 0009-0004-3591-5865
{Yinzhe~Shen}$^{2}$, %ORCID ID: 0009-0000-1956-4306
{Christian~Schlauch}$^{3}$, %ORCID ID: 0000-0002-6970-0094
{Mohamed-Khalil~Bouzidi}$^{3}$, % ORCID ID: 0009-0009-9734-3133
{Bojan~Derajic}$^{3}$, %ORCID iD: 0009-0000-6175-535X
{Natalie~Grabowsky}$^{4}$, % ORCID iD: 0009-0008-7944-6887
{Annajoyce~Mariani}$^{4}$, %ORCID iD: 0009-0001-6647-9996
{Dennis~Rochau}$^{4}$, % ORCID ID: 0009-0008-3717-9266
{Giovanni~Lucente}$^{5}$, %ORCID ID: 0000-0002-7844-853X
{Harsh~Yadav}$^{6}$,
{Firas~Mualla}$^{7}$,
{Adam~Molin}$^{8}$,
{Sebastian~Bernhard}$^{3}$, % ORCID-ID: 0000-0002-7194-7539
{Christian~Wirth}$^{3}$, % ORCID ID: 0009-0006-9755-6639
{{\"O}mer~{\c{S}}ahin~Ta{\c{s}}}$^{2}$, % ORCID ID: 0000-0002-1249-260X
{Nadja~Klein}$^{9}$, %ORCID ID: 0000-0002-5072-5347
{Fabian~B.~Flohr}$^{1}$,
and
{Hanno~Gottschalk}$^{4}$

\thanks{
$^{1}${Munich University of Applied Sciences, Intelligent Vehicles Lab (IVL), 80335 Munich, Germany (e-mail: intelligent-vehicles@hm.edu)}
$^{2}${FZI Research Center for Information Technology, 76131 Karlsruhe, Germany}
$^{3}${Continental Automotive Technologies GmbH, Guericke Str.7, 60488 Frankfurt a.\ M., Germany}
$^{4}${Technical University of Berlin, Institute of Mathematics, Straße des 17.Juni 136, 10587 Berlin, Germany}
$^{5}${German Aerospace Center (DLR), Institute of Transportation Systems, Lilienthalplatz 7, 38108 Braunschweig, Germany}
$^{6}${Aptiv Services Deutschland GmbH, Am Technologiepark 1, 42119 Wuppertal, Germany}
$^{7}${AI Lab, ZF Friedrichshafen AG, Uni-Campus D 5 1, 66123 Saarbruecken, Germany}
$^{8}${DENSO AUTOMOTIVE Deutschland GmbH, Freisinger Str.\ 21, 85386 Eching, Germany}
$^{9}${Karlsruhe Institute of Technology, Scientific Computing Center, Zirkel 2, 76131 Karlsruhe, Germany}
}
\thanks{The research leading to these results is funded by the German Federal Ministry for Economic Affairs and Energy within the project “NXT GEN AI METHODS – Generative Methoden für Perzeption, Prädiktion und Planung". The authors would like to thank the consortium for the successful cooperation.}
}

\markboth{Preprint}%
{Generative AI for Autonomous Driving: A Review}

\IEEEpubid{    }

\maketitle

\begin{strip}
  \centering
  \vspace{-45mm}

  \begingroup
  \def\svgwidth{\textwidth}
  \fontsize{6}{7}\selectfont
  %% Creator: Inkscape 1.4 (e7c3feb1, 2024-10-09), www.inkscape.org
%% Accompanies image file 'overview_links_v2.pdf' (pdf, eps, ps)
\begingroup%
  \makeatletter%
  \providecommand\color[2][]{%
    \errmessage{(Inkscape) Color is used for the text in Inkscape, but the package 'color.sty' is not loaded}%
    \renewcommand\color[2][]{}%
  }%
  \providecommand\transparent[1]{%
    \errmessage{(Inkscape) Transparency is used (non-zero) for the text in Inkscape, but the package 'transparent.sty' is not loaded}%
    \renewcommand\transparent[1]{}%
  }%
  \providecommand\rotatebox[2]{#2}%
  \newcommand*\fsize{\dimexpr\f@size pt\relax}%
  \newcommand*\lineheight[1]{\fontsize{\fsize}{#1\fsize}\selectfont}%
  \ifx\svgwidth\undefined%
    \setlength{\unitlength}{3437.25bp}%
    \ifx\svgscale\undefined%
      \relax%
    \else%
      \setlength{\unitlength}{\unitlength * \real{\svgscale}}%
    \fi%
  \else%
    \setlength{\unitlength}{\svgwidth}%
  \fi%
  \global\let\svgwidth\undefined%
  \global\let\svgscale\undefined%
  \makeatother%
  \begin{picture}(1,0.38184595)%
    \lineheight{1}%
    \setlength\tabcolsep{0pt}%
    \put(0,0){\includegraphics[width=\unitlength,page=1]{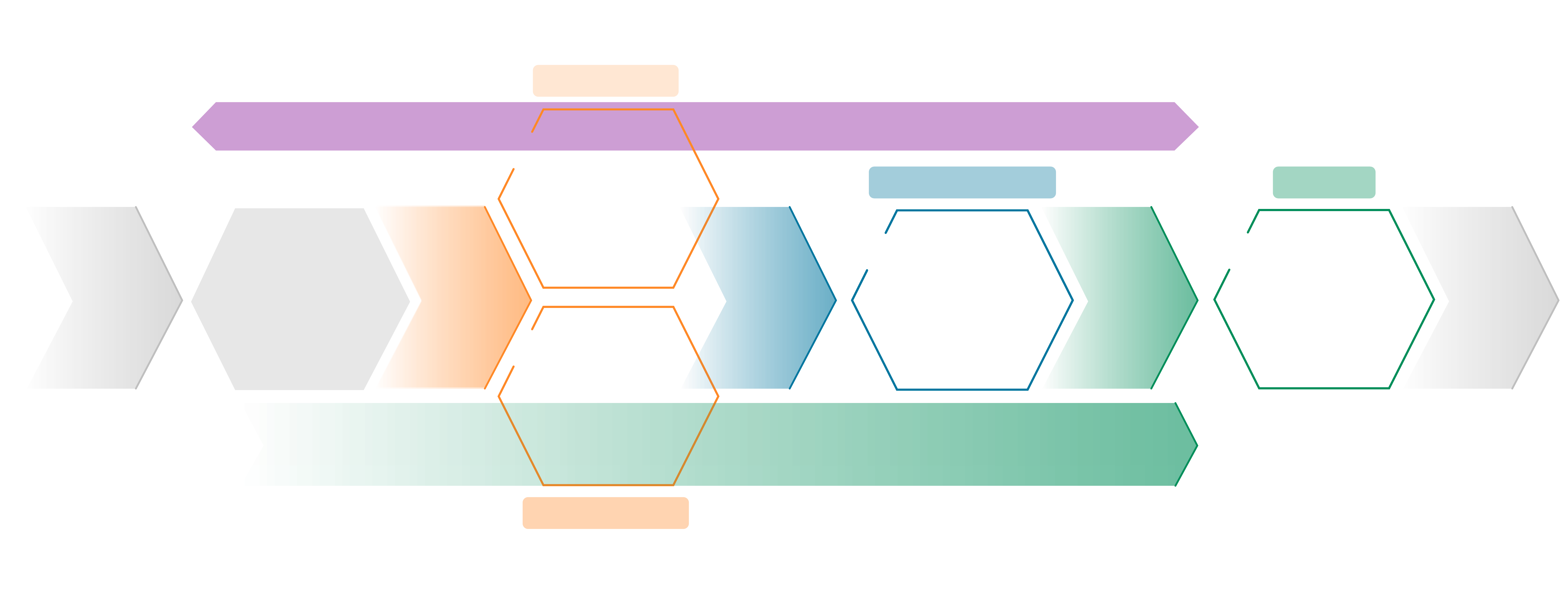}}%
    \put(0.05235489,0.1858872){\color[rgb]{0,0,0}\makebox(0,0)[lt]{\lineheight{1.25}\smash{\begin{tabular}[t]{l}Sensors\end{tabular}}}}%
    \put(0.34367627,0.32695848){\color[rgb]{0,0,0}\makebox(0,0)[lt]{\lineheight{1.25}\smash{\begin{tabular}[t]{l}Scene (abstract)\end{tabular}}}}%
    \put(0.34058254,0.05165766){\color[rgb]{0,0,0}\makebox(0,0)[lt]{\lineheight{1.25}\smash{\begin{tabular}[t]{l}Scene (abstract)\end{tabular}}}}%
    \put(0.46491518,0.20754413){\color[rgb]{0,0,0}\makebox(0,0)[lt]{\lineheight{1.25}\smash{\begin{tabular}[t]{l}Trajectory\\Prediction \&\\Forecasting\\Sec. \ref{sec:trajectory_prediction},\\p.~\pageref{sec:trajectory_prediction}\end{tabular}}}}%
    \put(0.69543252,0.19303333){\color[rgb]{0,0,0}\makebox(0,0)[lt]{\lineheight{1.25}\smash{\begin{tabular}[t]{l}Motion\\Planning\\Sec. \ref{sec:planning},\\ p.~\pageref{sec:planning}\end{tabular}}}}%
    \put(0.55562704,0.2629211){\color[rgb]{0,0,0}\makebox(0,0)[lt]{\lineheight{1.25}\smash{\begin{tabular}[t]{l}Expected Trajectories\end{tabular}}}}%
    \put(0,0){\includegraphics[width=\unitlength,page=2]{overview_links_v2.pdf}}%
    \put(0.81461235,0.26236983){\color[rgb]{0,0,0}\makebox(0,0)[lt]{\lineheight{1.25}\smash{\begin{tabular}[t]{l}Ego Motion\end{tabular}}}}%
    \put(0.92974184,0.18669885){\color[rgb]{0,0,0}\makebox(0,0)[lt]{\lineheight{1.25}\smash{\begin{tabular}[t]{l}Control\end{tabular}}}}%
    \put(0,0){\includegraphics[width=\unitlength,page=3]{overview_links_v2.pdf}}%
    \put(0.66759085,0.1001273){\color[rgb]{0,0,0}\makebox(0,0)[lt]{\lineheight{1.25}\smash{\begin{tabular}[t]{l}End-to-End\\Sec. \ref{sec:end-to-end}, p.~\pageref{sec:end-to-end}\end{tabular}}}}%
    \put(0.01467857,0.06301918){\color[rgb]{0,0,0}\makebox(0,0)[lt]{\lineheight{1.25}\smash{\begin{tabular}[t]{l}The AD-Stack\\Sec. \ref{sec:ad_stack}, p.~\pageref{sec:ad_stack}\end{tabular}}}}%
    \put(0,0){\includegraphics[width=\unitlength,page=4]{overview_links_v2.pdf}}%
    \put(0.27089189,0.20754413){\color[rgb]{0,0,0}\makebox(0,0)[lt]{\lineheight{1.25}\smash{\begin{tabular}[t]{l}Scene \&\\Scenario\\Generation\\Sec. \ref{sec:scene_generation},\\p.~\pageref{sec:scene_generation}\end{tabular}}}}%
    \put(0.14220093,0.18474551){\color[rgb]{0,0,0}\makebox(0,0)[lt]{\lineheight{1.25}\smash{\begin{tabular}[t]{l}Sensor Data (raw)\end{tabular}}}}%
    \put(0,0){\includegraphics[width=\unitlength,page=5]{overview_links_v2.pdf}}%
    \put(0.15915213,0.30528969){\color[rgb]{0,0,0}\makebox(0,0)[lt]{\lineheight{1.25}\smash{\begin{tabular}[t]{l}Datasets and Simulators\\Sec. \ref{sec:data}, p.~\pageref{sec:data}\end{tabular}}}}%
    \put(0.01399329,0.02463679){\color[rgb]{0,0,0}\makebox(0,0)[lt]{\lineheight{1.25}\smash{\begin{tabular}[t]{l}Fundamentals\\Sec. \ref{sec:fundamentals}, p.~\pageref{sec:fundamentals}\end{tabular}}}}%
    \put(0.01409046,0.36605944){\color[rgb]{0,0,0}\makebox(0,0)[lt]{\lineheight{1.25}\smash{\begin{tabular}[t]{l}Challenges, Recommendations and Outlook\\Sec. \ref{sec:discussion}, p.~\pageref{sec:discussion}\end{tabular}}}}%
  \end{picture}%
\endgroup%

  \endgroup

  \captionof{figure}[Visual overview of the survey structure]{Visual overview of the survey structure. The review begins with the Fundamentals (\Cref{sec:fundamentals}). \Cref{sec:ad_stack} introduces the general \acrfull{ad} stack architecture, followed by detailed examinations of its core components: Scene and Scenario Generation (\Cref{sec:scene_generation}), Trajectory Prediction and Forecasting (\Cref{sec:trajectory_prediction}), Motion Planning (\Cref{sec:planning}), and \acrfull{e2e} Driving (\Cref{sec:end-to-end}). The survey describes enabling Datasets and Simulators in \Cref{sec:data} and concludes with Challenges, Recommendations, and Outlook (\Cref{sec:discussion}).}
  \label{fig:ad_stack}
\end{strip}

%===============================================================================
% Abstract
%===============================================================================
\begin{abstract}
\Acrfull{genai} is rapidly advancing the field of \acrfull{ad}, extending beyond traditional applications in text, image, and video generation.
We explore how generative models can enhance automotive tasks, such as static map creation, dynamic scenario generation, trajectory forecasting, and vehicle motion planning.
By examining multiple generative approaches ranging from \acrfullpl{vae} and \acrfullpl{gan} to \acrfullpl{gt} and \acrfullpl{dm}, we highlight and compare their capabilities and limitations for \acrshort{ad}-specific applications.
Additionally, we discuss hybrid methods integrating conventional techniques with generative approaches, and emphasize their improved adaptability and robustness.
We also identify relevant datasets and outline open research questions to guide future developments in \acrshort{genai}.
Finally, we discuss five core challenges: safety, interpretability, realtime capabilities, training and benchmarking. We also present recommendations for image generation, dynamic scenario generation, and planning.
\end{abstract}

\begin{IEEEkeywords}
End-to-End Learning, Scene and Scenario Generation, Motion Planning, Trajectory Prediction
% Enter key words or phrases in alphabetical order, separated by commas.
% http://www.ieee.org/organizations/pubs/ani\_prod/keywrd98.txt
\end{IEEEkeywords}

%===============================================================================
% Venue and Deadlines
% IEEE Transactions on ITS 
%===============================================================================
% For both IEEE T-ITS and IEEE T-IV, page limitations are set as follows:
% "Survey Papers: 18 pages
% An author may submit manuscripts that exceed the suggested length but after the final acceptance,
% provided they pay an overlength charge of 175.00USD for each extra page."
% see https://ieee-itss.org/pub/t-its/#toc_Submission_Information_Information_for_Authors
% Second choice (if first submission fails): IEEE Transactions on Intelligent Vehicles (IV) 
%
% Venue proposals
% IEEE Access
% IEEE Transactions on Pattern Analysis and Machine Intelligence (TPAMI) - Survey category
% ACM Computing Surveys
% Springer Artificial Intelligence Review 
% Elsevier The Journal of Artificial Intelligence (AIJ)
% IEEE Transactions on ITS (+1 Fabian)
% IEEE Transactions on Intelligent Vehicles (IV) -- \textit{there are apparently ongoing problems -- is on hold} (+1 Fabian)
% Open Journal ITS (+1 Fabian)
% MDPI Robotics (Predatory!)
% Elsevier Transportation Research Part C: Emerging Technologies
% ACM Journal of Autonomous Transportation Systems
%===============================================================================
%===============================================================================

\section{Introduction}
\label{sec:introduction}
\iffalse
\begin{figure*}[htbp]
  \centering
  \def\svgwidth{\textwidth}
  \fontsize{6}{0}
  \input{Figures/hm/overview_links_v2.pdf_tex}
  \caption{Visual overview of the survey structure. The review begins with the Fundamentals (\Cref{sec:fundamentals}). \Cref{sec:ad_stack} introduces the general \acrfull{ad} stack architecture, followed by detailed examinations of its core components: Scene and Scenario Generation (\Cref{sec:scene_generation}), Trajectory Prediction and Forecasting (\Cref{sec:trajectory_prediction}), Motion Planning (\Cref{sec:planning}), and End-to-End driving (\Cref{sec:end-to-end}). The survey describes enabling Datasets and Simulators in \Cref{sec:data} and concludes with Challenges, Recommendations, and Outlook (\Cref{sec:discussion}).}
  \label{fig:ad_stack}
\end{figure*}
\fi

\Acrfull{genai} has shown significant success in the context of text, image, and video generation.
However, the principles of generative learning extend beyond these applications.
Generative learning describes algorithms that process data from a predefined source and learn to generate new data from a distribution that closely follows the source distribution.
In automotive applications, this capability can help to overcome limitations of existing approaches.

\acrfull{ad} systems fundamentally rely on the ability to perceive and understand their surroundings, and while large-scale street scene datasets \cite{cordts_cityscapes_2016,geiger_vision_2013,yu_bdd100k_2018} exist, they will never be near exhaustive. Therefore, \acrshort{genai} can be used to enrich existing datasets with rare or difficult-to-observe scenes. In addition to obvious use cases, such as generating new image and video sensor data, \acrshort{genai} can bridge the domain gap between synthetic and real data \cite{niemeijer_generalization_2024,schwonberg_survey_2023} and replace \cite{hu_gaia-1_2023, zhang_adding_2023} or improve \cite{mutze_semi-supervised_2023} conventional driving simulations.

\acrshort{genai} can capture not only the static context (e.g., roads and lanes) but also the dynamic context (e.g., vehicles, pedestrians, and cyclists) and model the temporal evolution of agents' intentions and movements. This allows for the transformation of scenes into dynamic scenarios.

Beyond scene and scenario generation, \acrshort{genai} also significantly contributes to trajectory forecasting.
Unlike conventional methods, \acrshort{genai}-based forecasting methods can sample from the distribution of possible agent behaviors, enabling the creation of diverse traffic simulations. Within the automotive domain, this is especially interesting because they can capture less likely but potentially safety-critical trajectories.

For similar reasons, \acrshort{genai} is also potentially helpful for (ego-)motion planning, as it is possible to create a diverse set of alternatives, including solutions for situations that require deviating from conventional driving habits.

\IEEEpubidadjcol

For all of the aforementioned applications, established methods such as \acrfull{il} for scenario generation and \acrfull{rl} for vehicle control already exist.
Thus, evaluating new generative solutions alongside these established methods is essential for assessing their relative strengths and limitations.
Such an analysis could facilitate the development of hybrid solutions that combine the strengths of individual methods.

While \acrshort{genai} opens many possible directions, the automotive domain imposes unique requirements that conventional \acrshort{genai} methods do not always meet.
Physical, regulatory, and cultural constraints require that generated data remain consistent with real-world requirements.
Equally important are safety concerns, as failures in assisted or \acrshort{ad} can lead to substantial costs or bodily harm.
Therefore, automotive \acrshort{genai} must also address rare but potentially hazardous scenarios.
Finally, automotive \acrshort{genai} applications often operate on low-performance systems such as embedded or edge devices.
As conventional \acrshort{genai} methods typically do not meet the consistency, safety or runtime requirements, specialized automotive-specific generative methods are required.

\subsection{Related Surveys and Selection Process}
\label{sec:intro_rw}
We provide a structured overview of \acrshort{genai} methods for \acrshort{ad}. \Cref{tab:surveys} lists existing surveys on \acrshort{genai} for \acrshort{ad}. Many surveys concentrate on specific applications, like scenario generation \cite{ding_survey_2023, yan_survey_2024, wang_survey_2024, gao_foundation_2025, song_generative_2025}, prediction \cite{dai_large_2025,wang_trends_2025} or planning \cite{li_large_2024}. Several reviews focus exclusively on foundation models, \cite{wu_prospective_2024,gao_survey_2024,gao_foundation_2025,dai_large_2025}, \acrfull{llm}-based methods \cite{dai_large_2025,li_large_2024,yang_llm4drive_2024} or world models \cite{wu_prospective_2024,guan_world_2024,feng_survey_2025}. In contrast, this survey systematically reviews \acrshort{genai} methods across the \acrshort{ad} stack, from scene and scenario generation through prediction and planning to \acrfull{e2e} frameworks. Wang et al.\ \cite{wang_generative_2025} focus on modality-specific generation and high-level applications, whereas this survey structures \acrshort{genai} methods by their role in the \acrshort{ad} stack, emphasizing model architectures, benchmarks and evaluation metrics. Figure~\ref{fig:ad_stack} illustrates the survey structure.

We selected the literature as follows. 
In the Fundamentals section, we cite breakthrough and seminal papers, while in the subject-specific sections we aim for comprehensiveness, referring to approximately 250 peer-reviewed papers selected through industry expertise. In the subject-specific sections, we privilege recent works in the last 6 years, up to March 2025, that were published in top-tier conferences and journals, including recent and highly-cited preprints. The publications were selected based on their application of \acrshort{genai} for \acrshort{ad}, following the  \acrshort{genai} taxonomy shown in Figure~\ref{fig:taxonomyGenL} and the AD application domains: scene and scenario generation, trajectory prediction and forecasting, motion planning, \acrshort{e2e} driving, and datasets and simulators.

\begin{table}[h!]
\centering
\caption{Overview of existing GenAI surveys. * indicates surveys published after submission. AD Stack includes SceneGen, ScenarioGen, Prediction, Planning, and related tasks.}
%\small
    \resizebox{0.95\columnwidth}{!}{%
    \begin{tabular}{@{}clll@{}}
\toprule
\textbf{Survey} & \textbf{Published} & \textbf{Application} & \textbf{Features} \\
\midrule
%A Survey on Safety-Critical Driving Scenario Generation - A Methodological Perspective
\cite{ding_survey_2023} & 02/2022 & ScenarioGen & Methods and tools for safety-critical scenario gen.\\

%A Survey of Generative AI for Intelligent Transportation Systems: Road Transportation Perspective
\cite{yan_survey_2024} & 12/2023 & ScenarioGen & Percept., pred., sim. and decision-making for traffic gen.\\

%A Survey of Scenario Generation for Automated Vehicle Testing and Validation
\cite{wang_survey_2024} & 12/2024 &ScenarioGen & Non-adaptive and adaptive scenario gen. for testing.\\

%Foundation Models in Autonomous Driving: A Survey on Scenario Generation and Scenario Analysis 
\cite{gao_foundation_2025}* & 06/2025 &ScenarioGen & \acrshort{llm}-, \acrshort{dm}-, world model- based Foundation models for scenario gen.\\

%Generative AI for Testing of Autonomous Driving Systems: A Survey 
\cite{song_generative_2025}* & 08/2025 & ScenarioGen & Distribution \acrshort{genai} models used in ADS testing.\\

%Large Foundation Models for Trajectory Prediction in Autonomous Driving: A Comprehensive Survey 
\cite{dai_large_2025}* & 09/2025 & Prediction & LLM-based Foundation models for trajectory prediction.\\

%Trends in Motion Prediction Toward Deployable and Generalizable Autonomy: A Revisit and Perspectives 
\cite{wang_trends_2025}* & 05/2025 &Prediction & Generalizability of motion prediction methods. \\

%Large Language Models for Human-like Autonomous Driving 
\cite{li_large_2024} & 07/2024 &Planning & LLM-based modular and \acrshort{e2e} approaches. \\

%LLM4Drive: A Survey of Large Language Models for Autonomous Driving 
\cite{yang_llm4drive_2024} & 11/2023 &AD Stack & \acrshort{llm}-based methods for \acrshort{ad} applications.\\

%Prospective Role of Foundation Models in Advancing Autonomous Vehicles 
\cite{wu_prospective_2024} & 12/2023 &AD Stack & \acrshort{llm}-, vision- and world model- based Foundation models for \acrshort{ad}.\\

%A Survey for Foundation Models in Autonomous Driving 
\cite{gao_survey_2024} & 02/2024 &AD Stack & Language, vision- and multi-modal Foundation models for \acrshort{ad}.\\

%World Models for Autonomous Driving: An Initial Survey
\cite{guan_world_2024} & 05/2024 &AD Stack & Evolution and SOTA of world models for \acrshort{ad}.\\

%Generative AI for Autonomous Driving: Frontiers and Opportunities 
\cite{wang_generative_2025}* & 05/2025 & AD Stack &  Reviews GenAI across modalities and reasoning. \\

%A Survey of World Models for Autonomous Driving 
\cite{feng_survey_2025}* & 09/2025 (v4) & AD Stack & World models for forecasting, behavior gen. and interaction.\\

\midrule
Ours & \textbf{--} & AD Stack & Unified taxonomy of GenAI across \acrshort{ad} stack; comprehensive SOTA \\
& & & analysis; identifies cross-category relationships, maps theoretical\\
& & & foundations to applications, and provides extensive discussion.\\
\bottomrule
\end{tabular}
\label{tab:surveys}
}
\end{table}

\section{Fundamentals} 
\label{sec:fundamentals}
Generative models are designed to approximate an unknown data-generating process.
This approximation process can be explicit, as in \acrfullpl{nf}, or implicit, as in \acrfullpl{gan}.
Generative models typically minimize one or more loss functions that measure how closely their generated distribution aligns with the true data distribution, typically in the form of divergence metrics like Kullback-Leibler or Jensen-Shannon \cite{cai_distances_2022}.
Directly minimizing these divergences can lead to instabilities like mode collapse,
where the model falls in a local minimum of the loss function, resulting in generated data with limited diversity.
To address these challenges, many modern \acrshort{genai} methods use auxiliary objectives. 
Additionally, divergence metrics often pose computational challenges.
To address this, surrogate metrics can evaluate model performance by focusing on quality and diversity rather than precise distance measurements.
Finally, classical learning frameworks, such as \acrfull{sl} for pre-training and \acrshort{rl} for optimizing long-term objectives, can be used to stabilize and guide generative model training.

\subsection{Methods}
\begin{figure*}
    \centering \includegraphics[width=\textwidth]{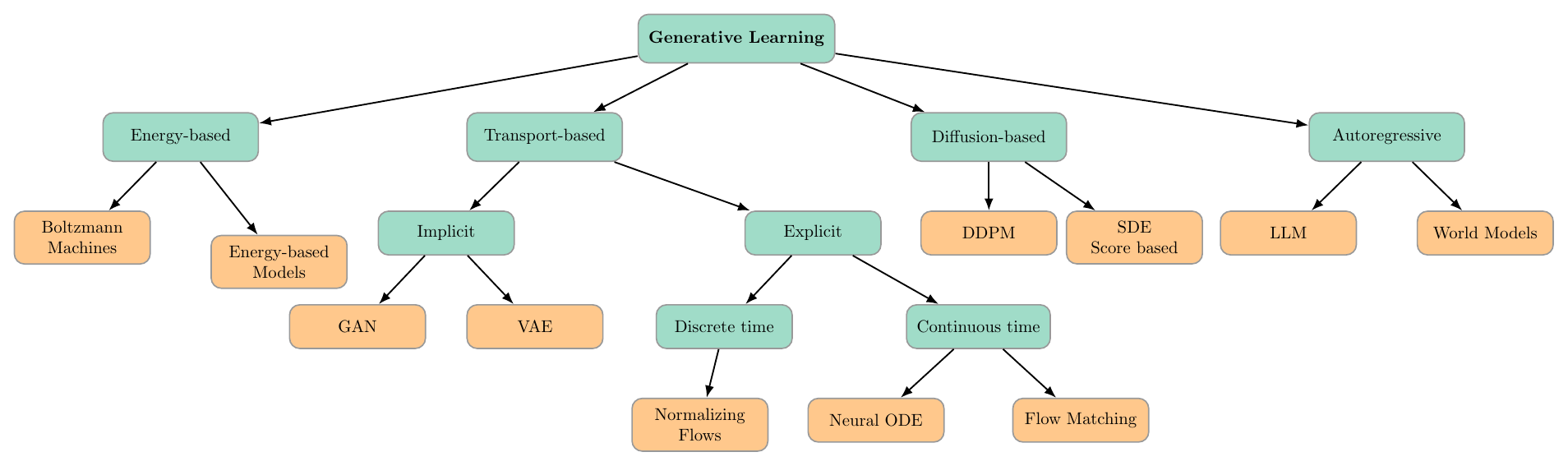}
    \caption{Taxonomy of generative learning approaches. Categories and subcategories  are shown in green, generative modeling approaches in orange. Hybrid forms may exist but are not listed. Sections \ref{sec:scene_generation} to \ref{sec:discussion} discuss applications of these methods to the AD stack.}
    \label{fig:taxonomyGenL}
\end{figure*}

We provide a brief overview of key models relevant to this survey, and categorize them as follows (see~\Cref{fig:taxonomyGenL}):
\paragraph{\acrfullpl{ebm}} follow the principles of thermodynamics and model the frequency of data proportional to the so-called Gibbs factor, which is the exponent of  a negative energy. In Boltzmann machines, energy is modeled as interaction energy of single bits, whereas for modern \acrshortpl{ebm} the energy function is represented by a neural network over a continuous feature space \cite{lecun_tutorial_2006,duvenaud_your_2020,carbone_hitchhikers_2024}. The densities given by \acrshortpl{ebm} are typically not normalized, and computing the normalizing factor is notoriously hard. Methods like Contrastive Divergence \cite{hinton_training_2002}, Score Matching \cite{hyvarinen_estimation_2005}, and Langevin Dynamics for sampling \cite{neal_mcmc_2011} address these difficulties. Also, sampling from \acrshortpl{ebm} requires Markov chain Monte Carlo methods.

\paragraph{Transport based models}directly learn a transport map from a source distribution to a target distribution that resembles the data distribution. The  map is usually encoded by a neural network and learned from the data. We distinguish between implicit and explicit models.

Implicit models like \acrfullpl{vae} \cite{kingma_auto-encoding_2022} or \acrshortpl{gan} \cite{goodfellow_generative_2020} often map noise from a lower latent dimension to the target distribution in higher dimensional data space. Due to the dimensional mismatch, these models do not possess a density and therefore cannot be trained likelihood based. \acrshortpl{vae} thus undergo a mixed training where a reconstruction loss between encoder and decoder is minimized, while the latent distribution is aligned with a normal distribution. \acrshortpl{gan} utilize an approximate variational representation of the Jensen-Shannon divergence with a discriminator network. Both methods are computationally efficient during generation and do not require specific network architectures.~\cite{croitoru_diffusion_2023}~\cite{saxena_generative_2022}

Explicit methods map a source noise distribution to a target data distribution, living in the same dimension as the target distribution. Therefore, the density of the generated distribution can be computed using the change of variables formula which includes inverses and determinants of the Jacobian matrix of the inverse. To make this computationally feasible, \acrshortpl{nf} \cite{kobyzev_normalizing_2021, rezende_variational_2015} directly model the inverse map and only invert the network once for generation. Discrete \acrshortpl{nf} achieve this using special architectures \cite{ishikawa_universal_2023}, whereas \acrlongpl{node} \cite{chen_neural_2018} conceptually perform a continuous morphing of the source distribution to the target distribution letting noise samples follow the flow of a vector field. The vector field is then modeled as a neural network that is not subject to architecture restrictions, as flow maps are always invertible. The likelihood can be easily computed from the trace of the Jacobian of the vector field. \acrfull{fm} \cite{lipman_flow_2023} avoids this procedure by directly regressing the vector field to conditional transport vector fields that are easy to compute.

\paragraph{\acrfullpl{dm}} model the target distribution as an invariant distribution of a stochastic process. This can be modeled either as a Markov chain in discrete time, like in classical \acrfullpl{ddpm} \cite{ho_denoising_2020} \cite{sohl-dickstein_deep_2015} \cite{song_score-based_2021}. The alternative is to use a continuous time stochastic process described by  a  (backward)  \acrfull{sde}. In both cases, the process in the normalizing direction is very simple, just adding noise to the data sample either in discrete portions or following an Ornstein-Uhlenbeck process in continuous time. In discrete time, the inverse denoising direction can be learned by estimating the noise vector in the present step, whereas the denoising direction in continuous time is known to follow a \acrshort{sde} with a drift that equals the score function, i.e. the derivative of the log-density of the data distribution. This can be exploited to directly learn the drift vector field. During inference, multiple steps are performed that reverse the noising process \cite{bansal_cold_2023, croitoru_diffusion_2023}.
\paragraph{Autoregressive models} are generative models that specifically generate data that unfolds in time, like human language or video data. As the time dimension adds another level of complexity to the data space, such models usually operate on a state space of reduced complexity. In the case of \acrshortpl{llm}, this is a representation of language as a sequence of tokens and the model learns a probabilistic prediction of the next token, on the basis of the previous token history. To achieve good performance on this task, the model must flexibly evaluate the prior history as context, which is achieved by the transformer architecture. \cite{vaswani_attention_2017}

Motivated by the success of \acrshortpl{llm}, world models use tokenizer maps, which are encoders to map a high dimensional state for the real world (or a video) to a lower dimensional latent representation. In the latent space, the  learning of the next `tokenized' state is performed mostly as a regression problem, in contrast to the classification problem on a discrete token space in the case of \acrshortpl{llm}. This coarse representation of the dynamics in the latent space is then mapped back to full states by learned super-resolution maps, that usually process a few consecutive frames to achieve temporal consistency of the rendering during generation. 

\begin{table}[t]
\centering
\caption{Properties of generative learning methods. The symbol ``--'' is used when the model class shows mixed performance w.r.t. the criterion. The markings show tendencies, but exceptions occur.}
\label{tab:GenAI_Properties}
\footnotesize
\setlength{\tabcolsep}{4.5pt}
\renewcommand{\arraystretch}{0.95}
\begin{tabular}{@{}lcccc@{}}
\toprule
\textbf{Model class} 
& \textbf{Distributional} 
& \textbf{Inference} 
& \textbf{Training} 
& \textbf{Time} \\
& \textbf{Accuracy} 
& \textbf{Speed} 
& \textbf{Stability} 
& \textbf{Series} \\
\midrule
Energy Based      & --       & $\times$    & $\times$    & $\times$ \\
Transport impl.   & --       & $\checkmark$ & $\times$   & -- \\
Transport expl.   & $\checkmark$ & --      & $\checkmark$ & -- \\
Diffusion         & $\checkmark$ & $\times$    & $\checkmark$ & -- \\
Autoregressive    & $\checkmark$ & -- & $\checkmark$ & $\checkmark$ \\
\bottomrule
\end{tabular}
\end{table}

In this review, we discuss generative models from the point of view of the underlying mathematical principles. While all models in \Cref{fig:taxonomyGenL} have the distributional universal approximation property, the machine learning community has common opinions on the strengths and weaknesses of the  respective model families. \Cref{tab:GenAI_Properties} provides an overview.
We further discuss the advantages and disadvantages of these paradigms in subsection \ref{subsec:advantages}. In practice, hybrid models are often used. For instance, latent diffusion combines a generative diffusion process with a \acrshort{vae} trained using a combined \acrshort{vae}-\acrshort{gan}-loss. \acrshortpl{ebm} also show promise in integration with other generative modeling frameworks, such as hybrid approaches combining energy functions with diffusion \cite{xu_energy-based_2024} or flow-based techniques \cite{chao_training_2024}. Hoover et al. \cite{hoover_energy_2024} show a combination of Transformers with energy-based learning.
Also, unconditional generation is rarely pursued but the output of generative models is controlled by external inputs, called prompts, which corresponds to learning to sample from target distributions conditioned on the prompt. A prominent model family accomplishing this are Vision Language Models (\acrshortpl{vlm}) where the prompt is an image that is fed into the decoding process of the language model via cross attention of the \acrshort{llm} to the visual tokens of a vision encoding model. Other examples relevant to this review include prompts that specify image or scene content, to which the generative model adapts its output.

\subsection{Classical learning strategies for generative models}
\label{sec:learning-strategies}
Classical learning strategies, i.e., \acrfull{usl}, \acrshort{sl},  \acrshort{rl}, and \acrshort{il}, mitigate training instabilities in generative models while improving divergence approximation efficiency and precision through auxiliary objectives. This section briefly overviews these techniques to support understanding of modern generative methods in \acrshort{ad}.

\paragraph{\acrfull{sl}}
Each sample in a \acrshort{sl} dataset consists of input features,
along with either a discrete or continuous label. The goal of a \acrshort{sl} task is to learn a function that maps input data to its corresponding label. 
A well-trained \acrshort{sl} model generalizes effectively to unseen data. However, models can suffer from two common issues: overfitting, where the model performs well on training data but poorly on new data due to poor generalization, and underfitting, where the model fails to learn meaningful patterns, resulting in poor performance on both training and new data. 
\acrshort{sl} is used to train generative models by providing explicit supervision for structured output generation. It serves as a pre-training mechanism for encoder-decoder architectures and aids in conditional generation tasks, such as text-to-image synthesis, as demonstrated in models like DALL·E.

\paragraph{\acrfull{usl}}
\acrshort{usl} involves the modeling of data without providing corresponding labels. 
The objective is to identify inherent structures, patterns, or distributions within the dataset. 
Success in \acrshort{usl} relies on extracting meaningful and generalizable insights, though it is inherently challenging to evaluate due to the absence of predefined labels. 
Variants of \acrshort{usl} include Self-Supervised Learning, which generates pseudo-labels from the data itself, often as a precursor to supervised tasks. Semi-Supervised Learning, by contrast, combines a small amount of labeled data with a larger set of unlabeled data, leveraging both to achieve improved performance when fully labeled datasets are scarce. These approaches bridge the gap between unsupervised and supervised paradigms, offering practical advantages in real-world applications with limited labeled data. 

\acrshort{usl} strategies are the foundation of deep generative models such as \acrshortpl{vae} and \acrshortpl{gan}, which self-organize features crucial for high-quality sample generation. Self-supervision extracts meaningful representations from sequential data, essential for training large-scale generative and foundation models.

\paragraph{\acrfull{rl}}
\acrshort{rl} \cite{sutton_reinforcement_2015, %szepesvari_algorithms_2010, 
%zhao_mathematical_2024
} is a machine learning framework in which an agent learns to take actions that maximize a reward signal while interacting with an unknown environment. RL tasks are typically modeled as \acrfullpl{mdp}, which define an agent’s interaction in a stochastic environment where the next state and reward depend only on the current state and action. An MDP consists of a state space representing information about the environment, an action space defining possible decisions, an unknown transition function governing state changes, and a reward function providing feedback for actions taken. 
A policy maps states to probability distributions over actions and defines the agent’s behavior. The goal of \acrshort{rl} is to find an optimal policy that maximizes the expected discounted cumulative reward, where future rewards are discounted to prioritize immediate rewards and stabilize learning.
\acrshort{rl} algorithms \cite{ghasemi_comprehensive_2025}
are categorized as model-based if they rely on an explicit model of the environment and model-free if they learn directly from interactions \cite{huang_model-based_2020}.
\acrshort{rl} refines generation policies via reward-based feedback, commonly used for fine-tuning language models and reinforcement-driven sampling.

\paragraph{\acrfull{il}}
\label{par:il}
Designing reward functions for optimal behavior in complex environments, such as \acrshort{ad}, is often impractical due to the diversity of scenarios. \acrshort{il} provides a viable alternative by training agents to mimic expert behavior directly from demonstrations \cite{zare_survey_2023}. This section highlights three main \acrshort{il} paradigms.
\acrfull{bc} employs \acrshort{sl} to map states to actions.
\acrshort{bc} can suffer from the covariate shift problem, where differences between training and testing distributions lead to compounding errors \cite{zare_survey_2023,codevilla_exploring_2019}. It is also prone to causal confusion, learning spurious correlations that result in irrational behaviors \cite{codevilla_exploring_2019}.
\acrfull{irl} infers the reward function of an agent by observing expert behavior, which is then used to optimize the agent’s policy via \acrshort{rl}. 
This ensures consistency between training and testing state distributions, addressing covariate shift \cite{zare_survey_2023}. However, \acrshort{irl} is computationally intensive and struggles with reward ambiguity \cite{arora_survey_2021}. 
\acrfull{aim} frames \acrshort{il} as a game between an agent and a discriminator, bypassing reward inference to reduce computational cost \cite{zare_survey_2023}. \acrshort{aim} excels in mimicking expert behavior but struggles with generalizing to unseen scenarios \cite{plebe_human-inspired_2024}.
\acrshort{il} variants enhance generative models by using expert demonstrations for structured generation, aiding controllable generation and refining multi-modal models.

\subsection{Conditioning, Prompting, Guidance}\label{sec:conditioning}
Unconditional generative models aim to generate samples from a marginal distribution $p(x)$ that most closely resembles the distribution of the training data. However, it is often more desirable to generate samples from specific subsets of the training data to enable targeted simulation \cite{guo_image_2021}. 
Conditional generative models introduce conditions to control the generation process. During inference, the model can generate samples from the conditional distributions $p(x| y)$ with respect to provided conditions $y$.

Conditions for predictive \acrshort{ad} systems are based on diverse modalities. Sensor-based models typically leverage sensor inputs like images and control actions like steering \cite{hu_gaia-1_2023, piazzoni_vista_2021, yang_generalized_2024}, while others rely on abstract representations of trajectories or agent dynamics as waypoints and traffic signals as constraining map elements \cite{zhong_guided_2023, jiang_motiondiffuser_2023, %huang_versatile_2024,
leonardis_genad_2025}. Alongside abstract representations, some methods condition the generation on handcrafted logic rules \cite{zhong_guided_2023} and cost functions \cite{jiang_motiondiffuser_2023, xu_diffscene_2023, huang_versatile_2024}. Frequently, text prompts provide context \cite{hu_gaia-1_2023} or driving instructions \cite{yang_generalized_2024, zhao_drivedreamer-2_2024}. 
Generative models are typically conditioned during the training process or in a subsequent fine-tuning stage.
\acrshort{ad} applications use diverse methods to leverage and combine approaches for conditional \acrshortpl{gan} \cite{bourou_gans_2024}, \acrshortpl{vae} \cite{ramchandran_learning_2024}, \acrshortpl{nf} \cite{winkler_learning_2019}, \acrshortpl{dm} \cite{jiang_motiondiffuser_2023, xu_diffscene_2023, huang_versatile_2024} and autoregressive models~\cite{seff_motionlm_2023}. 

A complementary approach for controlling the generation process can be achieved via online gradient-based guidance. Instead of introducing the conditioning mechanism during learning, this method guides the generation iteratively at the inference stage. 
This allows to define the desired behavior as an objective function \cite{jiang_motiondiffuser_2023, xu_diffscene_2023, huang_versatile_2024}, or a temporal logic formula \cite{zhong_guided_2023}, without the need of additional training.

\subsection{Modeling Uncertainty}
Modeling the data generating process involves two fundamental sources of uncertainty \cite{hullermeier_aleatoric_2021}. Aleatoric uncertainty arises from inherent ambiguities in the data, like measurement noise, whereas epistemic uncertainty stems from the model's inherent limitations. Quantifying these uncertainties is an essential part of \acrshort{ad} systems, since they affect the decision-making process of the \acrshort{ad} system \cite{bouzidi_motion_2024, mustafa_racp_2024}. Methods in uncertainty quantification typically focus on the compute-efficient estimation of the predictive uncertainty, following conformal prediction \cite{angelopoulos_gentle_2021} or Bayesian inference \cite{seligmann_beyond_2023} approaches. Recent methods also emphasize the robust disentanglement of uncertainty into aleatoric and epistemic components \cite{mucsanyi_benchmarking_2024}. %These methods include, for example, sampling-based techniques, such as deep ensembles~\cite{seligmann_beyond_2023}, \acrfull{mc} Dropout~\cite{gal_dropout_2016}, \acrlong{swag}~\cite{maddox_simple_2019}, Laplace~\cite{daxberger_laplace_2021} or variational approximations of Bayesian neural networks~\cite{loo_generalized_2021}, and deterministic uncertainty estimators~\cite{charpentier_training_2023, postels_practicality_2022}, such as spectral normalized Gaussian processes~\cite{liu_simple_2020}. 
In this context, generative models are used to explicitly model epistemic uncertainty as a distribution of latent variables \cite{goh_solving_2021,oberdiek_uqgan_2022,sun_conformal_2023}. 

Initial studies also stress the importance of tailoring uncertainty quantification to the generative modeling itself, e.g. to characterize the model's reliability. Consequently, methods for \acrshortpl{vae} \cite{bohm_uncertainty_2019}, \acrshortpl{dm} \cite{liu_ddm-lag_2024} and
\acrfullpl{gt} \cite{campos_conformal_2024, kim_adaptive_2024, ling_uncertainty_2024} are used to delineate whether hallucinations, inconsistencies or reproduction errors can be attributed to model limitations or inherent ambiguity in provided conditions.

\section{The Autonomous Driving Stack}
\label{sec:ad_stack}
\begin{figure*}[!t]
  \centering
  \begin{subfigure}[b]{0.3\textwidth}
    \centering
    \includegraphics[width=\textwidth, trim=0cm 1cm 1cm 4cm, clip]{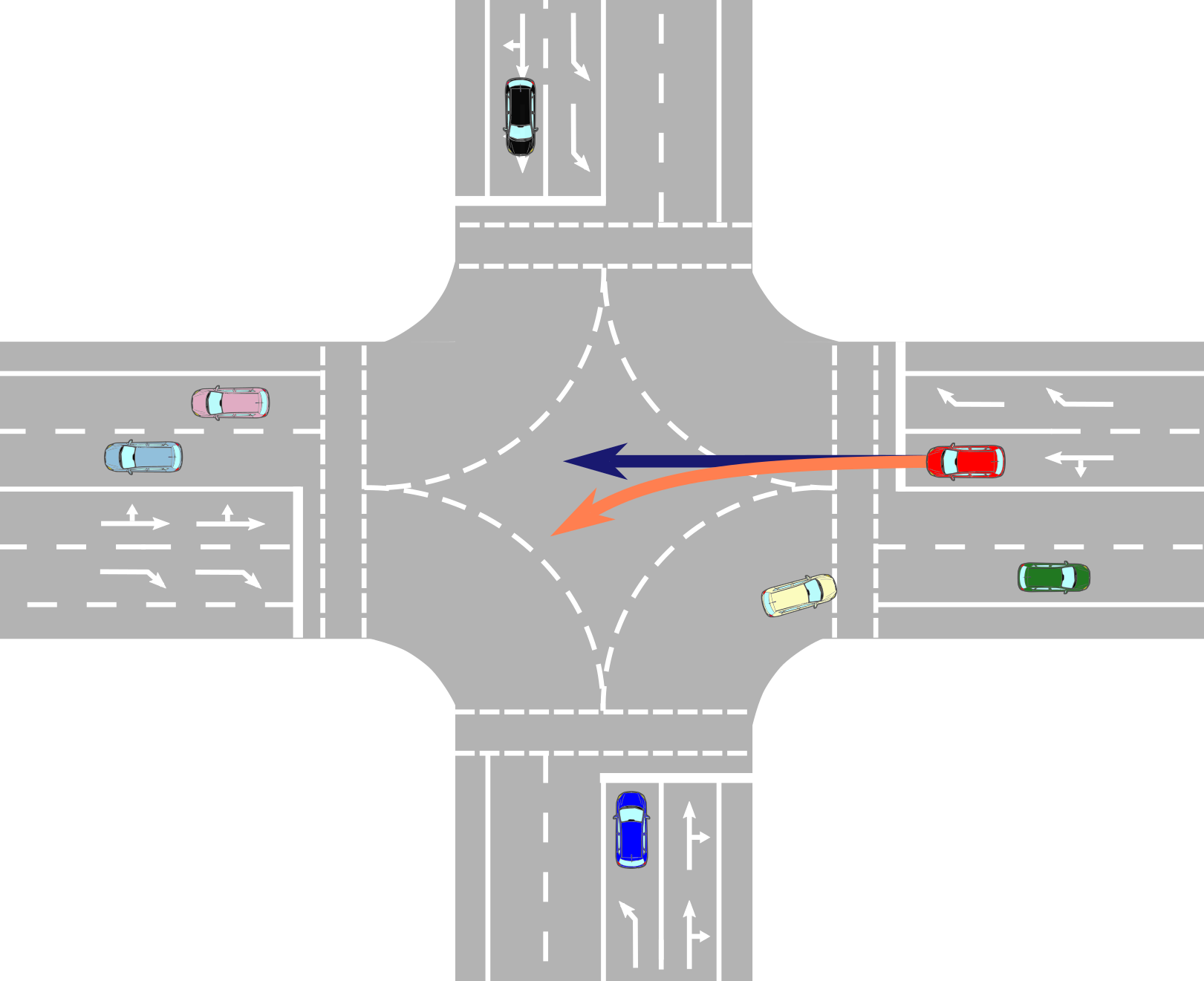}
    \caption{Marginal prediction}
    \label{fig:forecasting_marginal}
  \end{subfigure}
  \hfill
  \begin{subfigure}[b]{0.3\textwidth}
    \centering
    \includegraphics[width=\textwidth, trim=0cm 1cm 1cm 4cm, clip]{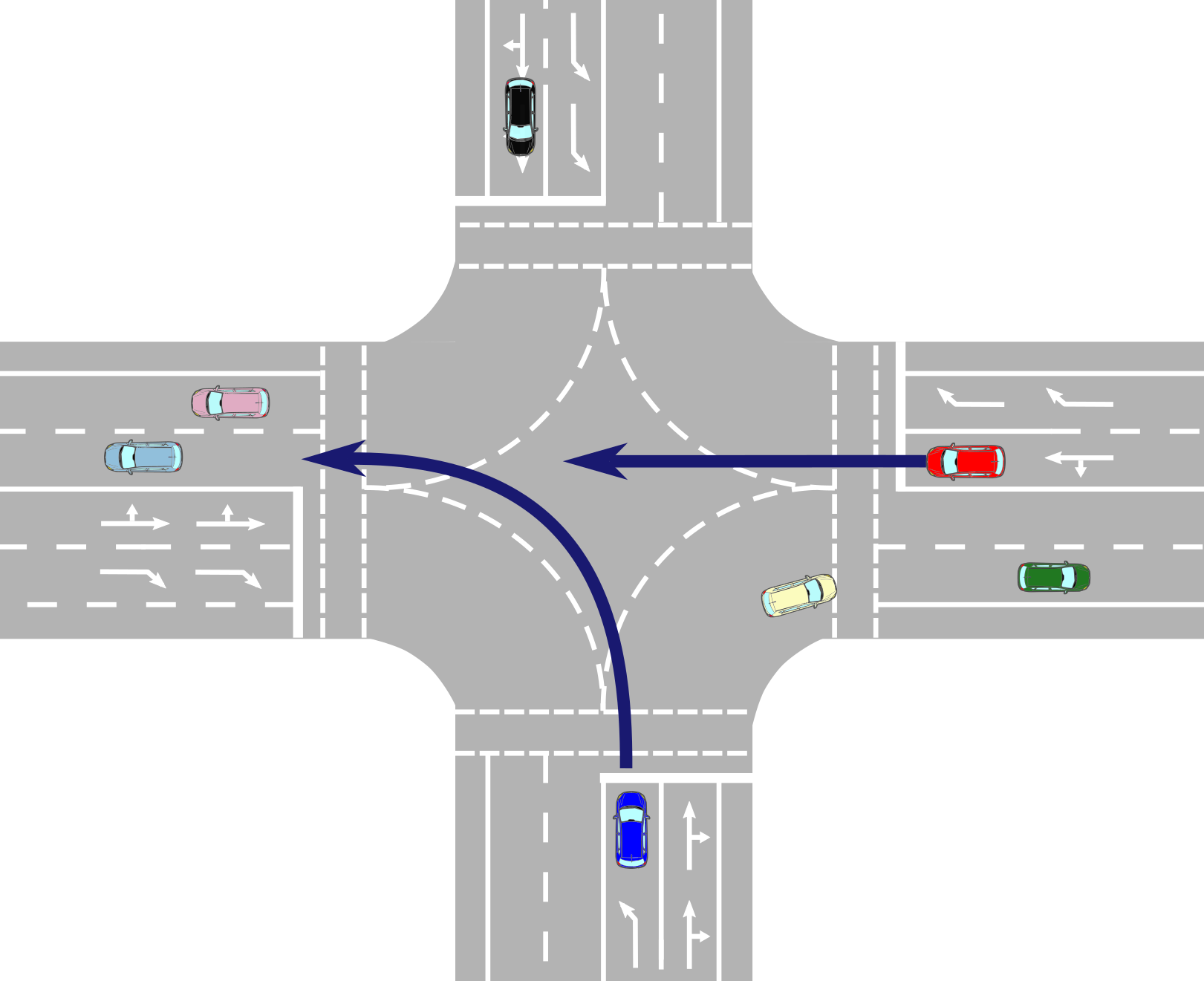}
    \caption{Conditional prediction}
    \label{fig:forecasting_conditional}
  \end{subfigure}
  \hfill
  \begin{subfigure}[b]{0.3\textwidth}
    \centering
    \includegraphics[width=\textwidth, trim=0cm 1cm 0cm 4cm, clip]{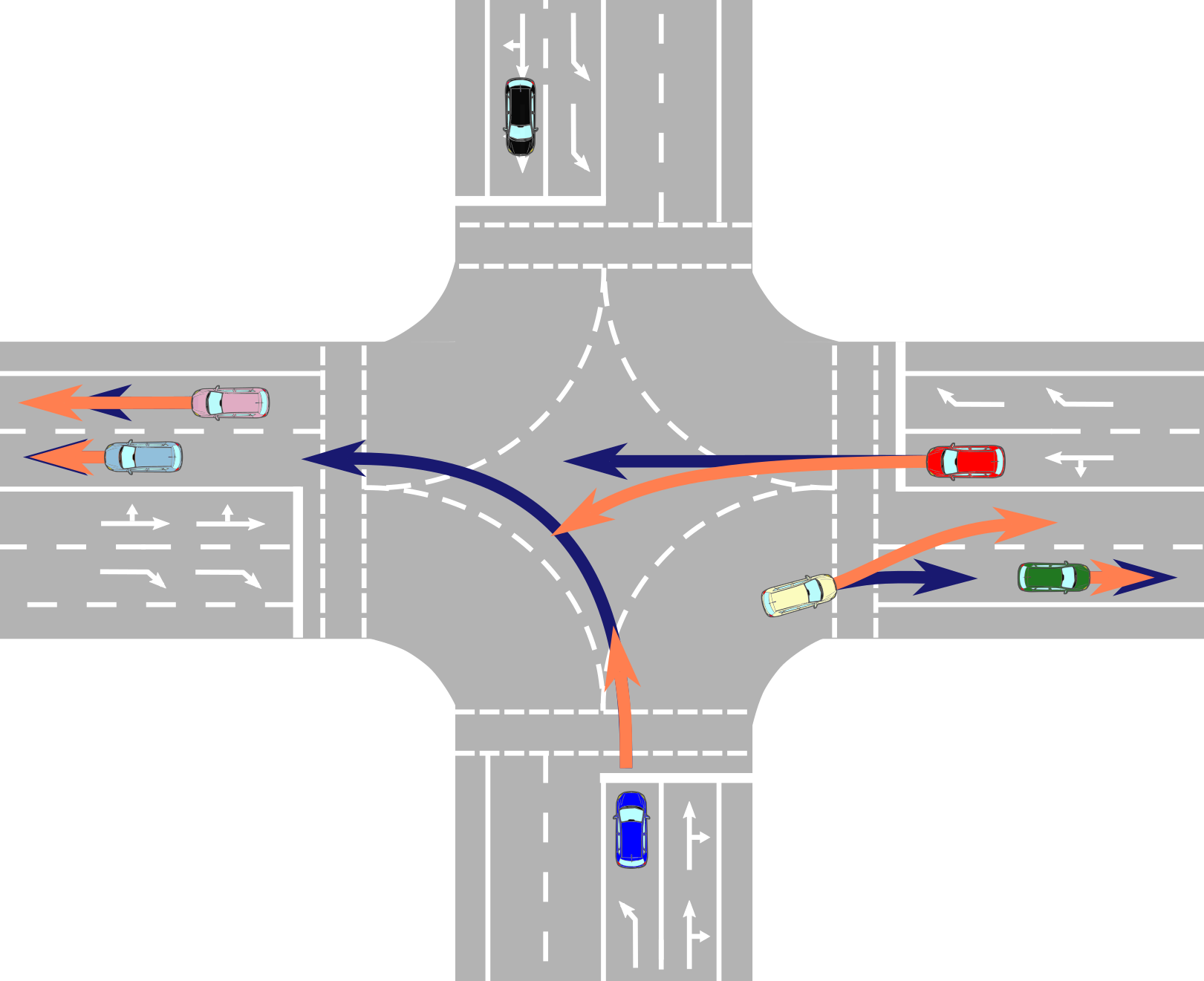}
    \caption{Joint prediction}
    \label{fig:forecasting_joint}
  \end{subfigure}
  \caption{Comparison of prediction types in a traffic intersection scenario. Blue and orange colors represent different prediction modes. The subfigure (a) shows marginal prediction for the red vehicle, (b) illustrates the conditional prediction of the red vehicle dependent on the ego vehicle (shown in blue), and (c) shows joint prediction, where blue and orange colors represent an entire scene independently.}
  \label{fig:forecasting_all}
\end{figure*}

\subsection{Components of an AD Stack}

A typical \acrshort{ad} stack consists of three modules (see~\Cref{fig:ad_stack}): perception, prediction and planning \cite{casas_mp3_2021, hu_st-p3_2022}. 
In addition to these modules, the AD stack relies on two hardware modules: sensors that collect raw data (cf.~\Cref{sec:input_modalities}), and actuators that execute control signals by adjusting throttle, brakes and steering of the vehicle.

\paragraph{Perception module}
Perception is about turning the raw sensor data into an intermediate representation of a higher abstraction level, such as occupancy grids and bounding boxes.
\acrshort{genai} can be used here to create virtual scenes or scenarios (\Cref{sec:scene_generation}), represented by either sensor information (e.g.\ images or videos) or directly as intermediate representation.
\paragraph{Prediction module} 
This module predicts the future states of traffic participants based on the intermediate representation.
This usually means forecasting trajectories, but autoregressive approaches using static scene forecast are also viable. 
\acrshort{genai} can serve as a proposal generator (\Cref{sec:trajectory_prediction}).
Additionally, a self-localization and/or mapping step may be integrated into prediction if the system relies on a map. 

\paragraph{Planning module} 
This module computes a feasible trajectory for the ego vehicle, based on the current scenario and the expected trajectories of other participants (\Cref{sec:planning}).
In addition to using \acrshort{genai} as a trajectory proposal generator for conventional planning methods, it is also possible to use \acrshort{genai} as a planner directly, considering the scenario and trajectories as a form of conditioning.

\paragraph{\acrlong{e2e} driving}
Instead of designing independent modules, it is also possible to  use a single system (\Cref{sec:end-to-end}). This means, the ego motion is directly computed, based on raw sensor data. This system can also be designed in a modular fashion, with a module for each sub task, but still as a network.

\subsection{Definition of Categories}
In the following, we shortly discuss some relevant dimensions to differentiate \acrshort{genai} approaches for \acrshort{ad}.

\subsubsection{Input Modalities}
\label{sec:input_modalities}

Generative models for \acrshort{ad} often rely on a variety of sensors to perceive the vehicle’s surroundings or an abstract representation of them. 
While RGB camera images are commonly used, combining them with other modalities -- such as LiDAR or radar point clouds -- provides robust, multimodal data for enhanced perception \cite{bai_transfusion_2022, yao_radar-camera_2024}. 
The integration of GPS coordinates and motion and orientation information from the Inertial Measurement Unit (IMU) can further enhance navigation and pose estimation \cite{kao_wei-wen_integration_1991, yurtsever_survey_2020}. 

Strategies to fuse these modalities are classified into early, late, or mid fusion. 
Late fusion processes input modalities separately before merging, enabling tailored processing for each. 
Early fusion concatenates raw inputs or extracted features into a joint representation, fostering cross-modal correlations. 
Mid fusion combines elements of both, merging modalities at an intermediate stage \cite{gadzicki_early_2020, xiao_multimodal_2022}.

Additional rich context from static and dynamic sources, such as \acrfull{hd}-Maps and occupancy grids, can be incorporated to further refine the model’s understanding of the environment. 
\acrshort{hd}-Maps provide precise geometric and semantic data to support vehicle localization, perception, and navigation \cite{bao_review_2023}. Annotated with attributes like lane topology, they offer rich static information on street networks and environments, often including instance embeddings \cite{dong_superfusion_2024, li_hdmapnet_2022}. Vectorized scene representations, unlike rasterized maps, enhance computational efficiency and include geometric street topology, instance-level details \cite{liao_maptr_2023, jiang_vad_2023, liu_vectormapnet_2022}, and dynamic objects \cite{liao_maptr_2023, gao_vectornet_2020}.
3D Occupancy Grids capture the volumetric state of the environment, classifying each voxel as free, unobserved, or occupied, optionally carrying semantic labels \cite{tian_occ3d_2023, wei_surroundocc_2023}.
\acrfull{bev} maps are a unified
top-down representation of the surrounding environment, partially addressing the challenges resulting from occlusions \cite{li_delving_2024}.
Natural language can serve as additional input for contextual guidance
enhancing interpretability and enabling flexible interactions with \acrshort{ad} systems \cite{wu_language_2023,cui_survey_2024}.
Past driving experiences from memory can extend the inputs in a temporal dimension.
The input data are mapped into a latent space directly or encoded into a format with reduced dimensionality by models like the \acrlong{vit} \cite{dosovitskiy_image_2021} for image-based data, PointNet \cite{qi_pointnet_2017} for LiDAR pointclouds, or the Q-Former \cite{li_blip-2_2023} for a joint representation of both image and text data.

\subsubsection{Static and Dynamic Contexts} 
In \acrshort{ad}, the surrounding context can be divided as static or dynamic. The static context comprises the surrounding elements which are static across time. Such elements include, for example, road maps, parked vehicles, lane boundaries, pedestrian crossings, construction sites and traffic signs. The dynamic context takes into account the scene elements that are not static and change their location or state. Road traffic including, for example, moving vehicles, pedestrians and bicycles, falls under this category. Traffic lights are also considered as dynamic contexts \cite{chai_multipath_2020,ngiam_scene_2021}.

\subsubsection{Marginal vs Conditional vs Joint Trajectory Forecasting}
\label{sec:marginal_conditional_joint_prediction}

The trajectory forecasting task (cf.~\Cref{sec:trajectory_prediction}) can be subdivided into three categories: Marginal, conditional, and joint, based on the interaction modeling of the agents in the scene. Marginal trajectory forecasting \cite{schafer_context-aware_2022,cui_survey_2024,chai_multipath_2020,%gao_vectornet_2020,
zhou_query-centric_2023} models the multi-modal trajectory distribution for every agent in the scene individually, conditioned on the static and dynamic context information in the previous time steps. In marginal trajectory forecasting, future interactions among neighboring agents are ignored. Conditional trajectory forecasting relaxes this assumption by modeling the target agent's future trajectory given fixed trajectories of the neighboring agents \cite{sun_m2i_2022,tolstaya_identifying_2021}. However, this one-way dependency assumption still fails in highly dynamic traffic situations, where neighboring agents mutually influence each other's motion. In such cases, joint trajectory forecasting is necessary to model the interaction among the future multi-modal trajectories of all the agents in a scene \cite{park_leveraging_2023, luo_jfp_2023, wagner_jointmotion_2024, wagner_redmotion_2024}. However, the complexity of the joint distribution over all agents’ future multi-modal trajectories grows combinatorially with the number of agents. This motivates the development of compute-efficient approaches to interaction modeling (see~\Cref{sec:trajectory_prediction}). The idea of marginal, conditional and joint trajectory forecasting is shown in \Cref{fig:forecasting_all}.

\section{Scene and Scenario Generation}
\label{sec:scene_generation}
\begin{table*}[h!]
\centering
\caption{Overview of scene and scenario generation methods, with focus on static maps (S), dynamic scenarios (D) and world models (W)}
\label{tab:scene-scenario-models}
\resizebox{1.0\textwidth}{!}{%
\begin{tabular}{@{}lccccc@{}}\toprule
\textbf{Model} & \textbf{Year} & \textbf{Approach} & \textbf{Category} & \textbf{Benchmarks} & \textbf{Features} \\\midrule
Neural Turtle Graphics \cite{chu_neural_2019} & 2019 & Autoregressive (encoder-decoder \acrshort{rnn}) & S & RoadNet/SpaceNet & Generates plausible road network graphs \\
HDMapGen \cite{mi_hdmapgen_2021} & 2021 & Autoregressive (graph) & S & Argoverse & Hierarchical graph model for lane-level HD maps \\
SLEDGE \cite{chitta_sledge_2024} & 2024 & Diffusion (latent) & S & nuPlan & Generates lane graph and agents; integrated traffic simulation \\
DriveSceneGen \cite{sun_drivescenegen_2024} & 2024 & Diffusion (image) + graph & S & Waymo & Two-stage: rasterized scene generation and agent simulation \\
PolyDiffuse \cite{chen_polydiffuse_2023} & 2023 & Diffusion (set) & S & nuScenes/Argoverse2 & Guided set diffusion for HD map reconstruction from sensor data \\
TrafficGen \cite{feng_trafficgen_2023} & 2023 & Autoregressive (encoder-decoder) & D & Waymo & Autoregressive vehicle and trajectory generation\\
RealGen \cite{ding_realgen_2025} & 2024 & Transport (retrieval autoencoder) & D & nuScenes & Retrieval based controllable traffic scenarios \\
LCTGen \cite{tan_language_2023} & 2023 & Transport (LLM conditioned) & D & Waymo & Language-conditioned traffic scene generation \\
World Models \cite{ha_world_2018} & 2018 & Transport (VAE) + autoregressive \acrshort{rnn} & W & CarRacing-v0/VizDoom TakeCover & Trains agent in hallucinated latent environment \\
PlaNet \cite{hafner_learning_2019} & 2019 & Transport (latent RSSM) + CEM & W & DeepMind Control Suite & Latent dynamics and planning from pixels \\
DreamerV3 \cite{hafner_mastering_2024} & 2024 & Transport (latent world model) & W & Atari/Atari100k/ProcGen/DMLab/Minecraft/DMControl/BSuite & General world model across 150+ tasks with fixed hyperparameters \\
GAIA-1 \cite{hu_gaia-1_2023} & 2023 & Autoregressive transformer + diffusion (video) & W & Proprietary UK urban driving & Multimodal generative world model for controllable driving videos \\
DriveDreamer-4D \cite{zhao_drivedreamer4d_2024} & 2024 & Diffusion (video) + 4D Gaussian Splatting & W & Waymo/nuScenes/PandaSet & World model priors for 4DGS novel trajectory views \\
ADriver-I \cite{jia_adriver-i_2023} & 2023 & Diffusion (video) + autoregressive control & W & nuScenes/private highway & Joint planning and future scene generation \\
Think2Drive \cite{li_think2drive_2024} & 2024 & Transport (latent world model) & W & CARLA Leaderboard v2/CornerCaseRepo & Model-based RL planner for quasi-realistic CARLA v2 corner cases \\
Drive-WM \cite{wang_driving_2023} & 2023 & Diffusion (multiview video) & W & nuScenes & Multiple futures generation for planning safety assessment \\
LAW \cite{li_enhancing_2024} & 2025 & Transport (latent self-supervision) & W & nuScenes/NAVSIM/CARLA & Self-supervised latent future prediction for \acrshort{e2e} driving \\
BEVWorld \cite{zhang_bevworld_2024} & 2024 & Diffusion (latent BEV) & W & nuScenes/CARLA & Multimodal BEV world model for image and lidar forecasting \\
MUVO \cite{bogdoll_muvo_2024} & 2025 & Transport (multimodal latent world model) & W & CARLA & RGB–LiDAR world model with 3D occupancy prediction \\
OccWorld \cite{zheng_occworld_2023} & 2023 & Autoregressive occupancy transformer & W & Occ3D/nuScenes & 3D occupancy world model for 4D scene and ego forecasting \\
Drive-OccWorld \cite{yang_driving_2025} & 2025 & Autoregressive 4D occupancy world model & W & nuScenes/Lyft-Level5 & 4D occupancy forecasting and planning with action conditions \\
OccLLaMA \cite{wei_occllama_2024} & 2024 & Autoregressive occupancy-language-action model & W & Occ3D/nuScenes/nuScenes-QA & VLA occupancy world model for forecasting, planning and VQA \\

\bottomrule
\end{tabular}}
\end{table*}

Scene and scenario generation is a key enabler for developing and validating \acrshort{ad} systems. The size of scenes varies significantly, ranging from larger environments such as virtual cities to the immediate surroundings of an ego vehicle. Modeling road users' intent and motion is critical for transforming static environments into realistic, time-evolving scenarios. A scenario describes a temporal sequence of scenes, each scene being a snapshot of the environment at a point in time. Every scenario begins with an initial scene, and its progression can be guided by specified actions, events, goals, and values. This temporal aspect distinguishes a scenario, which spans a period of time, from a scene, which describes a single point in time or a limited snapshot \cite{ulbrich_defining_2015}. This section covers three facets of generative scene and scenario creation: generating detailed static maps of road networks and infrastructure, modeling temporal, dynamic multi-agent interactions with realistic behavior, and forecasting world models that integrate static and dynamic elements to predict how the environment will evolve. An overview of representative methods in these areas, including their approaches and key features, is summarized in \Cref{tab:scene-scenario-models}.

\subsection{Static Map Generation}
Static maps, or \acrshort{hd}-Maps \cite{liu_high_2020, elghazaly_high-definition_2023}, provide the foundational structural layer for \acrshort{ad} systems, including road geometry and traffic rules. While typically generated via diverse sensor suites (LiDAR, camera, radar, aerial imagery) or manual annotation \cite{yang_robust_2018, tang_high-definition_2023, alghooneh_unified_2024, wei_creating_2022, elhousni_automatic_2020}, these approaches are costly. Recent research focuses on dynamic, real-time map generation \cite{liao_maptr_2023, zhang_online_2023}. Generative models in this domain address road layout generation, \acrshort{hd}-Map generation, and sensor-conditioned generation.
Static maps are integral in creating realistic city and simulation environments for testing and validation of \acrshort{ad} functions. Such simulations rely on these maps to replicate real-world conditions. While these maps need to be plausible and sufficiently detailed in a data-driven manner, they do not need to correspond to actual locations in the real world. This differs significantly from \acrshort{hd}-Maps used by \acrshort{ad} systems, which are required to precisely represent real-world conditions.

\paragraph{Road Topology} Road topology generation involves creating structured representations of road networks, capturing their layout and connectivity. Some works extract road topologies from aerial images \cite{mattyus_deeproadmapper_2017}, \cite{li_topological_2019}, but these approaches are constrained by the spatial resolution of aerial images and cannot generate new topologies conditioned on existing city styles. To overcome these limitations, Chu et al \cite{chu_neural_2019} propose Neural Turtle Graphics, a sequential generative model that iteratively creates new nodes and edges that are connected on a graph to generate a plausible city road layout. This method is used to generate natural road layouts based on existing or completely novel city styles. 
\paragraph{HD-Map Synthesis} Whereas previous methods focus on generating plausible road layouts, other methods extend this to the generation of detailed \acrshort{hd}-Maps. 
In \cite{mi_hdmapgen_2021}, the authors introduce HDMapGen, a set of autoregressive models that generate high-quality and diverse \acrshort{hd}-Maps in a data-driven manner. They explore sequence-based, plain graph, and hierarchical graph representations, finding that the hierarchical approach best preserves the natural structure of \acrshort{hd}-Maps. This two-level model includes a global and a local graph decoder, employing recurrent Graph Attention Networks and multi-layer perceptrons to produce two separate graphs. Evaluation focuses on topology fidelity, geometry fidelity, urban planning features, and diversity.
The method of generating lane graphs from \acrshort{hd}-Maps can also be directly integrated into \acrshort{e2e} driving simulators. For example, SLEDGE \cite{chitta_sledge_2024} generates lane graphs and improves on the quality and scalability of HDMapGen. In addition, agents are generated jointly and in parallel to directly simulate traffic behavior on the generated lane graph. 
SLEDGE rasterizes the scene, encodes it with a raster to vector autoencoder, generates or inpaints maps via latent diffusion \cite{lugmayr_repaint_2022}, then simulates agents with the Intelligent Driver Model \cite{treiber_congested_2000}.
Similarly, DriveSceneGen \cite{sun_drivescenegen_2024} generates and simulates both static map elements and dynamic traffic participants in a two-stage model. It uses an image-space \acrshort{dm} to generate a rasterized \acrshort{bev} representation of a scene. Following, a graph-based vectorization algorithm transforms the representation into a vector format. In the second stage, agents are simulated and multi-modal behaviors are predicted with the Motion Transformer Model \cite{shi_motion_2022}.

\paragraph{Sensor-Conditioned Reconstruction} Diffusion Methods have been successfully employed in generative modeling of diverse domains like images, video and LiDAR pointclouds. Techniques like SLEDGE or DriveSceneGen rasterize scenes to leverage established diffusion techniques in 2D image space and demonstrate that generative models can be used to generate new static scenes for simulation purposes. PolyDiffuse \cite{chen_polydiffuse_2023} shows that \acrshortpl{dm} can also be used for a structured reconstruction task of set elements like polylines and polygons. Using a Guided Set \acrshort{dm}, vectorized \acrshort{hd}-Maps are created through a generation process conditioned on visual sensor data. Guidance networks are introduced to control the noise injection to prevent the permutation of the set elements from getting lost in the diffusion process. In the standard diffusion process, different permutations of a sample gradually become indistinguishable as the process progresses. Therefore, specific noise is injected at the element level with Guidance Networks. The reverse diffusion process denoises an initial proposal which is either from a human annotator or a task-specific method like MapTR \cite{liao_maptr_2023}, \cite{liao_maptrv2_2024} to reconstruct and refine a \acrshort{hd}-Map conditioned on visual sensor data.

The development of generative models for static map generation has significantly advanced beyond traditional methods.
However, the field remains in its early stages, with limited models addressing the diverse challenges of \acrshort{hd}-Map generation for both simulation and real-time \acrshort{ad} applications.

\subsection{Generation of Temporal and Dynamic Aspects}
Recently, various methods have been developed to generate and modify the dynamic aspects of driving scenarios, offering different levels of control and realism.

\begin{figure}[htbp]
    \centering
    \begin{subfigure}[t]{0.24\textwidth}
        \centering
        \includegraphics[width=\linewidth]{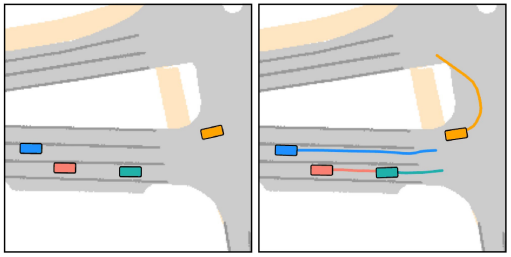}
        \caption{Initial position: left-hand side image. Tag: \textit{U-Turn}}
    \end{subfigure}
    \hfill
    \begin{subfigure}[t]{0.24\textwidth}
        \centering
        \includegraphics[width=\linewidth]{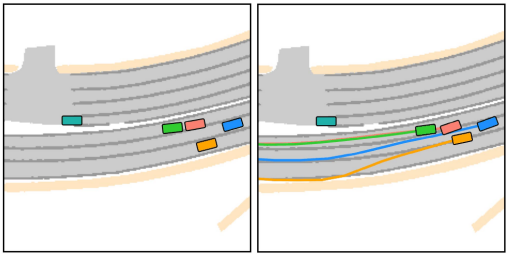}
        \caption{Initial position: left-hand side image. Tag: \textit{Lane change}}
    \end{subfigure}
    \hfill
    \caption{Demonstration of scenario generation starting from an initial position and a tag. Images were taken from \cite{ding_realgen_2025}.}
    \label{fig:realgen}
\end{figure}

\begin{figure}[htbp]
    \centering
        \includegraphics[width=0.75\linewidth, trim=1.3cm 4cm 2cm 3cm, clip]{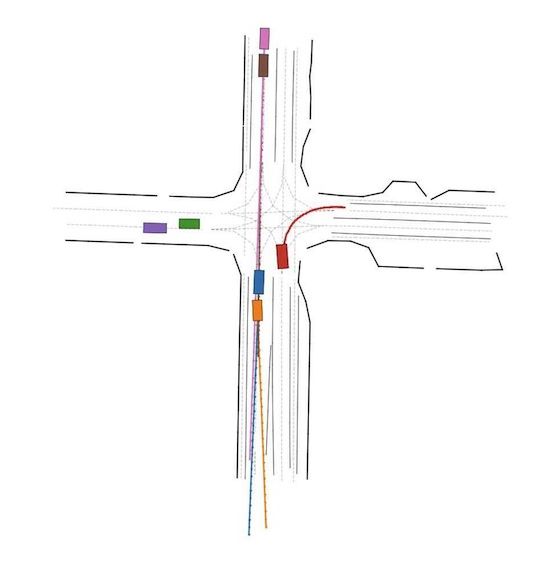}
        \caption{Demonstration of text-prompt to scenario generation. Image was taken from \cite{tan_language_2023}. Prompt: \textit{The scene is very dense. There are only vehicles on the left side of the center car. Most cars are moving at high speeds. The ego-vehicle turns right.}}
    \label{fig:lctgen}
\end{figure}

TrafficGen \cite{feng_trafficgen_2023} uses an encoder-decoder to create or modify scenarios in vectorized format. This enables the creation of new scenarios and the modification of existing ones by adding new agents or modifying the trajectories of current ones. However, this method has limitations, as the user lacks control over the quality or specific properties of the generated scenarios, which is a common issue with auto-encoding of training data. In this respect, RealGen \cite{ding_realgen_2025} (cf.~\Cref{fig:realgen}) goes one step further by generating scenarios that closely resemble a query scenario (e.g., U-turn or lane change). Additionally, it employs \acrlong{rag}, ensuring that the generated scenario is not only similar to the query but also to the \acrlong{knn} of the query in external datasets, which were not seen during training. The highest level of control over scenario generation is achieved with approaches like LCTGen \cite{tan_language_2023}. Here, users can generate a new scenario (cf.~\Cref{fig:lctgen}) or modify an existing one using a text prompt. The crux of the approach is leveraging \acrshortpl{llm} through advanced prompt engineering to convert textual descriptions into a representation where each agent and the map are described by a few parameters. Specifically, each agent is parameterized by pose relative to the ego vehicle and future actions. The map is parameterized by lane count and lane positions relative to the ego vehicle.  

\subsection{World Models}
\label{subsec:world-models}
Maps and bounding boxes are handcrafted abstractions of driving scenarios. Comprehensive scenario generation requires world models.
Rooted in psychology~\cite{craik_nature_1967, tolman_cognitive_1948}, control theory~\cite{bryson_applied_2018}, and model-based \acrshort{rl}~\cite{sutton_dyna_1991}, the concept was introduced for training AI agents by Ha and Schmidhuber~\cite{ha_world_2018}.

Despite diverse applications~\cite{hu_gaia-1_2023, zheng_occworld_2023, wang_driving_2023}, the definition of world models remains unclear. 
Generally, a world model is capable of 1) representing the environment (real or virtual) in a latent space, and 2) forecasting future states based on history and conditions (e.g., ego actions, text).
LeCun~\cite{lecun_path_nodate} further argues that world models should grasp common-sense physics, such as object permanence, beyond simple prediction.

Early world model studies~\cite{ha_world_2018, hafner_dream_2020, hafner_learning_2019} focus on \acrshort{rl} in simulations and video games, typically using \acrlong{cnn}-based autoencoders for environmental representation. The seminal work~\cite{ha_world_2018} employs a \acrfull{rnn} to predict the next latent state based on actions, enabling recursive forecasting without real-world interaction, while incorporating a Mixture Density Network to model uncertainty. PlaNet~\cite{hafner_learning_2019} introduces the recurrent state-space model (RSSM) to separate stochastic and deterministic components, improving long-term prediction. The Dreamer series~\cite{hafner_dream_2020, 
%hafner_mastering_2022, 
hafner_mastering_2024} advances these models, training action and value functions in the latent space with high sample efficiency.

In \acrshort{ad}, observations can be multi-modal, e.g., images, point clouds, text, and annotations. \mbox{GAIA-1}~\cite{hu_gaia-1_2023} learns from images and text with \acrshort{usl}.
It suggests that high-level semantics, rather than low-level textures, are important for world modeling. It distills an image tokenizer from DINO~\cite{caron_emerging_2021} to ensure a meaningful image representation. The image tokens can further fuse with text and action tokens and predict the future autoregressively using Transformers. 
A concurrent work DriveDreamer~\cite{leonardis_drivedreamer_2025} additionally uses annotated bounding boxes and HDmaps as inputs, which reduces the amount of training data required. Followups improve DriveDreamer in multi-view video generation, scene manipulation, and reconstruction~\cite{zhao_drivedreamer-2_2024, zhao_drivedreamer4d_2024}. 
ADriver-I~\cite{jia_adriver-i_2023} adopts concept to that of GAIA-1, but trains a multi-modal \acrshort{llm} on a private dataset to encode the image and text inputs. Unlike GAIA-1, it outputs control signals together with future videos, enabling motion planning.
Think2Drive~\cite{li_think2drive_2024} applies the model-based \acrshort{rl} method DreamerV3~\cite{hafner_mastering_2024} to \acrshort{ad} in the simulator CARLA~\cite{dosovitskiy_carla_2017}. \mbox{Drive-WM~\cite{wang_driving_2023}} employs world models to forecast the future with diverse ego maneuvers, resulting in a decision tree. It samples the action with the highest reward, which aims to keep a safe distance to other objects and drive within lanes. \Cref{sec:planning} provides more planning related literature. 

Beyond image-based visual fidelity, alternative representations improve spatial reasoning and efficiency. LAW~\cite{li_enhancing_2024} supervises the world model in the latent space. Similar to JEPA~\cite{lecun_path_nodate}, LAW predicts the next latent state of the world and compares it with the real embedding at the same timestep. Some other works~\cite{zhang_bevworld_2024, bogdoll_muvo_2024} represent the latent world in a 2D space to facilitate learning spatial information, where BEVWorld~\cite{zhang_bevworld_2024} predicts multiple timesteps in parallel instead of autoregression.
3D occupancy is another representation of the world, which simplifies the world and retain only semantic and geometric information. Training occupancy world models require less learning effort than image ones, e.g., OccWorld~\cite{zheng_occworld_2023} and Drive-OccWorld~\cite{yang_driving_2025} are trained with 8 GPUs on the nuScenes dataset~\cite{caesar_nuscenes_2020}. In contrast, image world models require 32 or even more GPUs, and larger dataset (usually private) than nuScenes. Although occupancy can be combined with other input modalities, such as text~\cite{wei_occllama_2024}, the fidelity of an occupancy world model relies on the quality of the conversion from sensor data to occupancy voxels.

World models open new possibilities for \acrshort{ad}, including generating realistic training data, discovering rare scenarios and corner cases, and providing a unified approach to prediction and planning. 
However, key challenges remain, such as assessing the validity of generated scenarios for training \acrshort{ad} models, measuring the accuracy of the modeled world, and improving the data efficiency of training world models.

\subsection{Scene and Scenario Generation Metrics}

Generative scene and scenario models are commonly evaluated in terms of fidelity, realism, and diversity. 
For static maps, existing metrics mainly assess distributional similarity and reconstruction quality. 
Distributional metrics compare generated maps with real road networks using graph, geometry, and urban-planning statistics, such as lane connectivity, Laplacian spectrum, segment length, orientation, density, reachability, and map diversity \cite{chu_neural_2019,mi_hdmapgen_2021}. 
Reconstruction metrics instead compare generated and ground-truth lane graphs at the instance level, for example through point matching, topology-aware matching, lateral offset, Chamfer distance, and AP/mAP under distance thresholds \cite{liao_maptr_2023,chen_polydiffuse_2023,chitta_sledge_2024}. 

\begin{table}[t]
\centering
\caption{Metrics for static scene generation}
\label{tab:static-scene-metrics}
\small
\resizebox{\columnwidth}{!}{%
\begin{tabular}{@{}lll@{}}
\toprule
\textbf{Category} & \textbf{Metric} & \textbf{Description} \\
\midrule
\multirow{5}{*}{\textbf{Distribution}} 
& Topology fidelity (MMD) \cite{mi_hdmapgen_2021} & Graph degree, Laplacian stats \\
& Geometry fidelity (Fr\'echet) \cite{mi_hdmapgen_2021} & Lane length and orientation stats \\
& Urban planning (Fr\'echet) \cite{chu_neural_2019} \cite{mi_hdmapgen_2021} & Density, connectivity, reach, convenience \\
& Map diversity (Chamfer) \cite{chu_neural_2019} \cite{mi_hdmapgen_2021} & Novelty and internal diversity of graphs \\
& Perceptual FID \cite{heusel_gans_2017} \cite{chu_neural_2019} & Inception features on rendered maps \\
\midrule
\multirow{7}{*}{\textbf{Reconstruction}} 
& GEO metric \cite{chitta_sledge_2024} & Point match precision, recall, F1 \\
& TOPO metric \cite{chitta_sledge_2024} & GEO on local reachability subgraphs \\
& Directed GEO / TOPO \cite{chitta_sledge_2024} & GEO or TOPO with direction constraint \\
& Lateral L2 error \cite{chitta_sledge_2024} & Mean lateral offset of matched points \\
& Chamfer distance (points) \cite{liao_maptr_2023} \cite{chitta_sledge_2024} & Bidirectional distance of sampled graphs \\
& AP / mAP (Chamfer) \cite{liao_maptr_2023} \cite{chitta_sledge_2024} & Detection AP with Chamfer thresholds \\
& Order aware angle metric \cite{liao_maptr_2023} \cite{chen_polydiffuse_2023} & Chamfer plus average angle distance \\
\bottomrule
\end{tabular}}
\end{table}

While these metrics capture important aspects of map plausibility, diversity, and reconstruction quality, they do not by themselves establish downstream utility. 
In particular, the literature provides limited evidence that distributional static-map metrics correlate reliably with downstream AD performance. 
This is a relevant limitation because prediction and planning modules may be sensitive to local and task-specific map errors, for example missing lane connectivity, wrong lane direction, topological breaks at intersections, or inaccurate drivable-area boundaries. 
Adjacent work therefore increasingly complements map-level evaluation with downstream task evaluation: predicted vectorized HD maps have been fed into motion forecasting models \cite{liu_vectormapnet_2023}, online map uncertainty has been propagated to trajectory prediction \cite{gu_mapuncertainty_2024}, enriched map topology has been evaluated through map-based motion prediction \cite{naumann_Lanelet2NuScenes_2023}, and generated lane graphs have been used in planner-in-the-loop simulation \cite{chitta_sledge_2024}. 
Future work on static map generation should therefore report predictor- or planner-in-the-loop metrics in addition to the distributional and reconstruction metrics.

\section{Trajectory Forecasting}
\label{sec:trajectory_prediction}
\begin{figure*}[!t]
  \centering
  \begin{minipage}{0.48\linewidth}
  \centering
      \fbox{
  \begin{subfigure}[b]{0.3\textwidth}
    \centering
    \includegraphics[width=\textwidth, trim=0cm 0cm 0cm 0cm, clip]{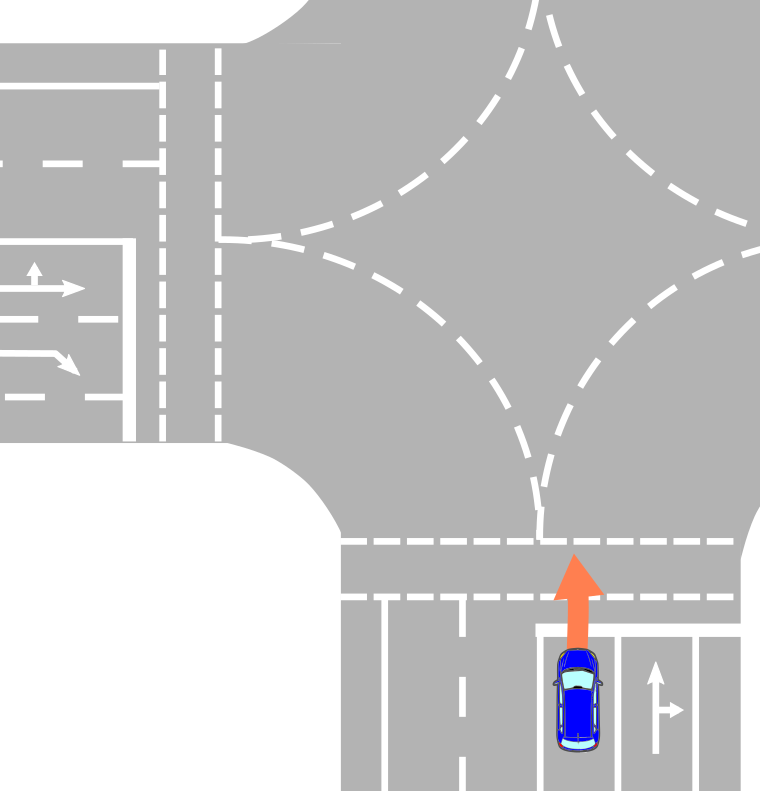}
    \caption{Step 1\ldots}
    \label{fig:generation_t1}
  \end{subfigure}
  \begin{subfigure}[b]{0.3\textwidth}
    \centering
    \includegraphics[width=\textwidth, trim=0cm 0cm 0cm 0cm, clip]{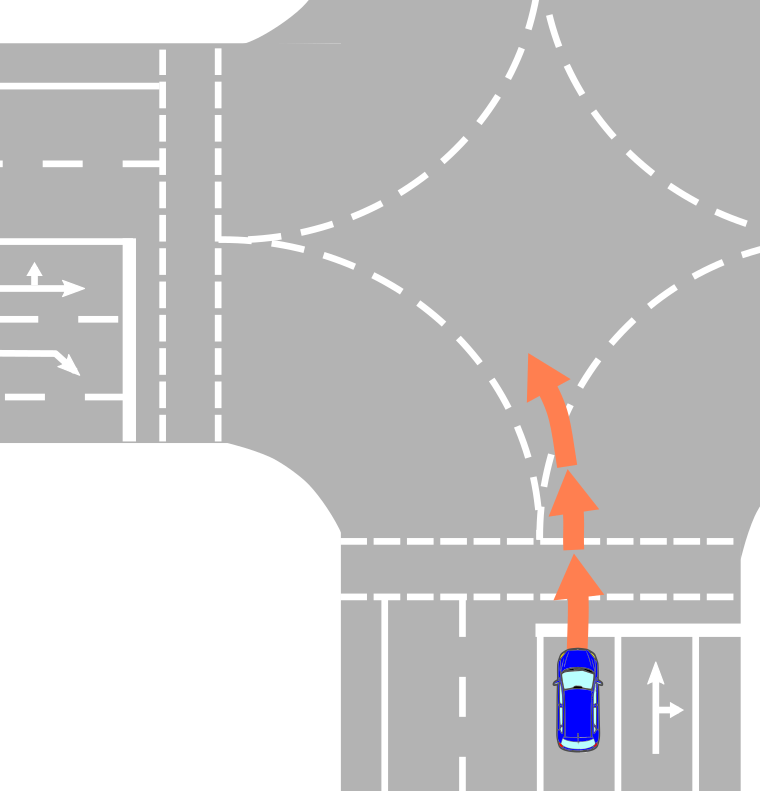}
    \caption{Step 3\ldots}
    \label{fig:generation_t3}
  \end{subfigure}
  \begin{subfigure}[b]{0.3\textwidth}
    \centering
    \includegraphics[width=\textwidth, trim=0cm 0cm 0cm 0cm, clip]{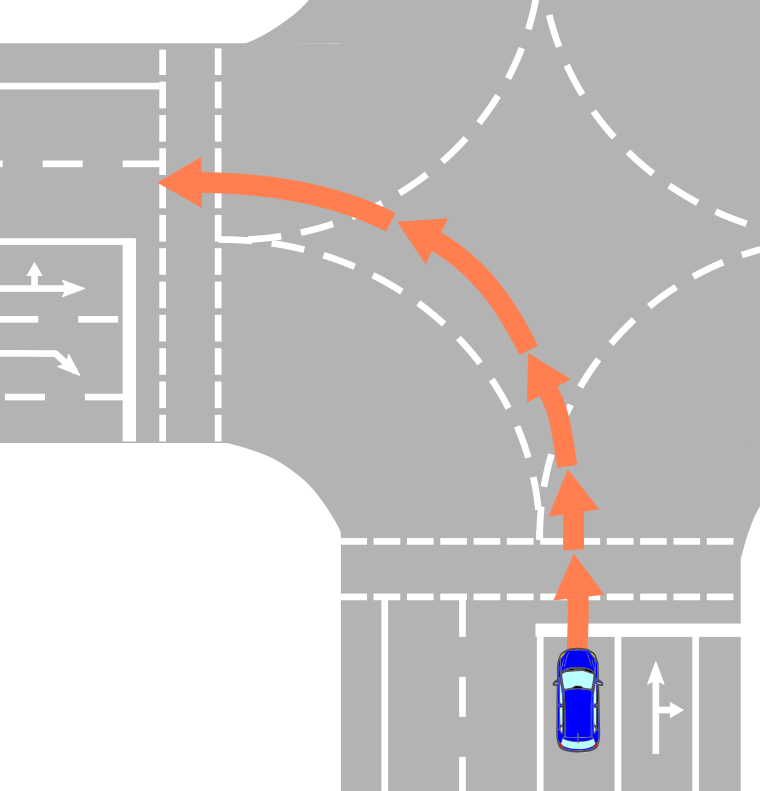}
    \caption{Step 5\ldots}
    \label{fig:generation_t5}
  \end{subfigure}}
  \caption{Autoregressive generation iteratively progresses motion trajectories over $1, 2, \ldots, N$ timesteps.}
  \label{fig:generation_autoregressive}
\end{minipage}
\begin{minipage}{0.01\linewidth}
\end{minipage}
 \begin{minipage}{0.48\linewidth}
 \centering
       \fbox{
  \begin{subfigure}[b]{0.3\textwidth}
    \centering
    \includegraphics[width=\textwidth, trim=0cm 0cm 0cm 0cm, clip]{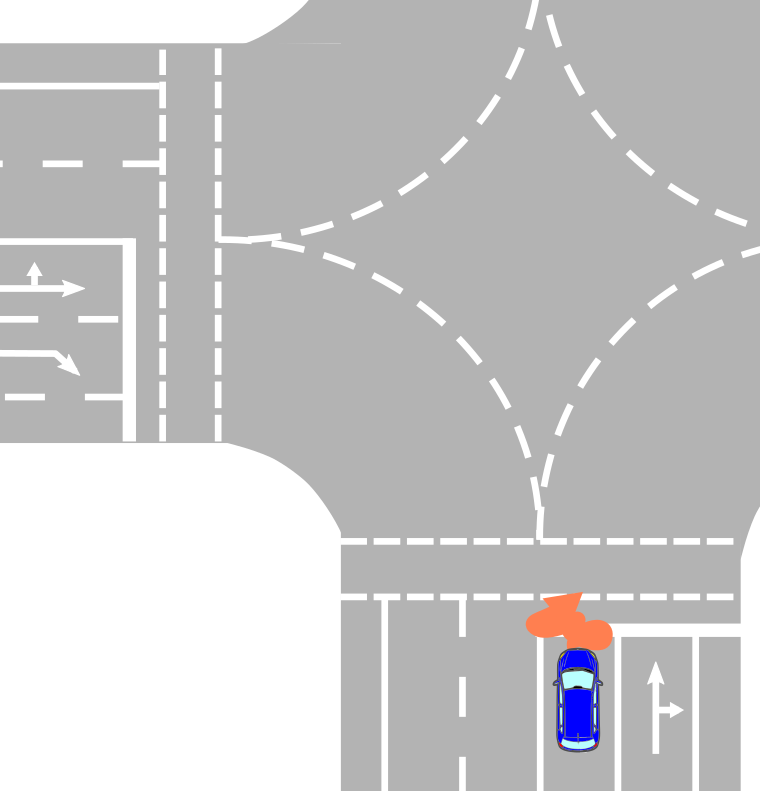}
    \caption{Step 10\ldots}
    \label{fig:generation_step10}
  \end{subfigure}
  \begin{subfigure}[b]{0.3\textwidth}
    \centering
    \includegraphics[width=\textwidth, trim=0cm 0cm 0cm 0cm, clip]{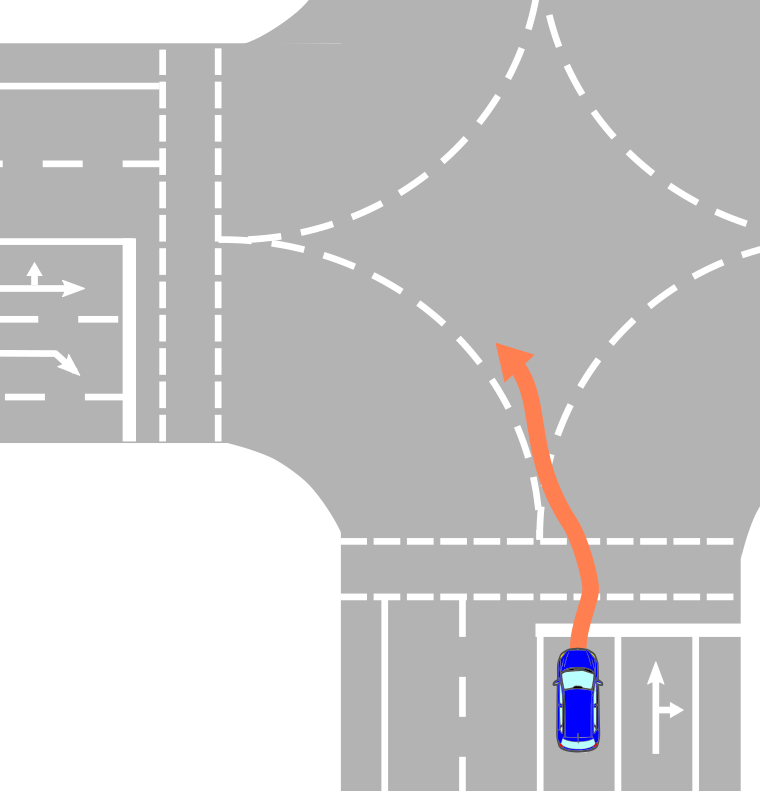}
    \caption{Step 100\ldots}
    \label{fig:generation_step100}
  \end{subfigure}
  \begin{subfigure}[b]{0.3\textwidth}
    \centering
    \includegraphics[width=\textwidth, trim=0cm 0cm 0cm 0cm, clip]{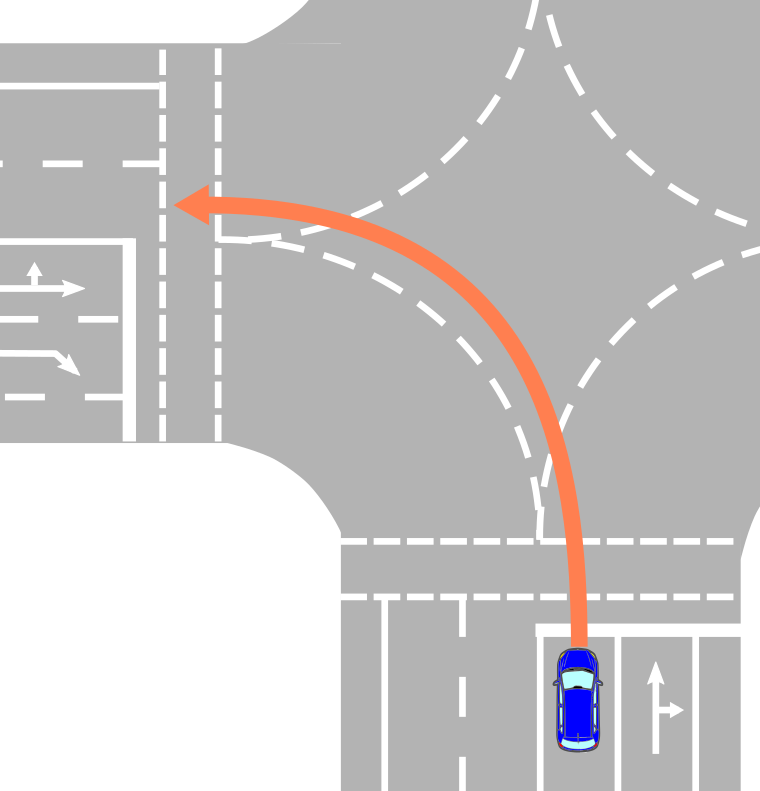}
    \caption{Step 500\ldots}
    \label{fig:generation_step500}
  \end{subfigure}}
  % \vspace{-60px}
  \caption{Diffusion-based generation iteratively refines motion trajectories over $1, 2, \ldots, N$ denoising steps.}
  \label{fig:generation_diffusion}
  \end{minipage}
\end{figure*}

The actions of a safe \acrshort{ad} system's actions depend on the predicted future states of the surrounding agents. Tractably encoding interactions between these agents as context remains a fundamental challenge for all trajectory forecasting models (see~\Cref{sec:marginal_conditional_joint_prediction} for details). %Recently applied generative decoding strategies, , allow for new levels of control and variation in the trajectory forecasting process. 
In this section, we explore recent context encoding and generative decoding strategies, such as autoregressive (see~\Cref{fig:generation_autoregressive}) and diffusion-based approaches (see~\Cref{fig:generation_diffusion}). Selected methods are summarized in Table~\ref{tab:trajectory-forecasting-models}.

\subsection{Context Encoding}

Generative trajectory forecasting architectures leverage the same encoding strategies as their discriminative counterparts, largely adopting attention-based mechanisms to model multi-agent interactions and context relations. Full self-attention mechanisms model multi-agent interactions explicitly along both spatial and temporal dimensions but scale exponentially with agent count and time horizon \cite{nayakanti_wayformer_2023}. This is especially problematic in joint trajectory forecasting (see ~\Cref{sec:marginal_conditional_joint_prediction}) when considering interactions over future time steps. Factorized attention mechanisms provide a low-cost approximation by independently processing temporal and spatial dimensions \cite{ngiam_scene_2021, nayakanti_wayformer_2023}. Latent variable models \cite{girgis_latent_2022, luo_jfp_2023, nayakanti_wayformer_2023} compress queries into shared latent representations instead, providing an amortized approximation of the joint behavior. Hierarchical and local attention \cite{zhou_hivt_2022, shi_motion_2022, wagner_redmotion_2024} decomposes the shared environment into spatially local regions. Similarly,  graph-based models \cite{cui_lookout_2021, luo_jfp_2023, zhou_query-centric_2023} constrain connectivity through neighborhood graphs or k-nearest agents, achieving near-linear scaling. Other methods limit the temporal resolution, attending only to interactions at specific time steps or at a lower frequency over agent trajectories \cite{wu_smart_2024}.

Many approaches within these groups fuse the multi-agent interaction encoding with other context information, such as the road topology, environmental conditions or agent intentions,  using multi-head cross-attention. While this context information is often provided in the form of grids, graphs, or goal points, Xia et al. \cite{xia_language-driven_2024} provide an example where multi-head cross-attention is used to fuse map and interaction features with language-derived vehicle and scene descriptors.

\subsection{Generative Trajectory Decoding}

\begin{table*}[h!]
\centering
\caption{Overview of Generative Trajectory Forecasting Methods}
\label{tab:trajectory-forecasting-models}
\resizebox{1.0\textwidth}{!}{%
{
\begin{tabular}{@{}lcccccc@{}}\toprule
\textbf{Model} & \textbf{Year} & \textbf{Approach} & \textbf{Interaction Type} & \textbf{Benchmarks} & \textbf{Features} \\\midrule

MFP \cite{tang_multiple_2019} 
  & 2019 
  & Autoregressive 
  & joint 
  & NGSIM, Argoverse 
  & parallel per-agent \acrshortpl{rnn} with shared latent variable attention\\ 

LB-EBM \cite{pang_trajectory_2021}  
  & 2021 
  & EBM  
  & joint 
  & Stanford Drone Dataset, ETH-UCY 
  & inverse reinforcement learning with latent beliefs\\

SEEM \cite{wang_seem_2022}  
  & 2022 
  & EBM  
  & joint 
  & ETH-UCY 
  & generator and energy network architecture\\

EPD \cite{liu_modeling_2024}  
  & 2022 
  & EBM \& Diffusion  
  & joint 
  & ETH-UCY 
  & hybrid energy and diffusion prediction model\\

LatentFormer \cite{abolfathi_latentformer_2022} 
  & 2022 
  & Autoregressive
  & joint 
  & nuScenes 
  & transformer encoder-decoder conditioned on learned latent intention variables \\

MotionLM \cite{seff_motionlm_2023} 
  & 2023 
  & Autoregressive
  & joint 
  & Waymo, Argoverse~2 
  & transformer encoder-decoder predicting over discrete heuristic motion token vocabulary\\

Trajeglish \cite{philion_trajeglish_2024} 
  & 2023 
  & Autoregressive 
  & joint 
  & Waymo, nuScenes 
  & transformer encoder-decoder with data-driven motion tokenization \\ 

AMP \cite{jia_amp_2024} 
  & 2024 
  & Autoregressive
  & joint 
  & Waymo
  & decoder-only transformer with motion and map tokenization \\ 

SMART \cite{wu_smart_2024} 
  & 2024 
  & Autoregressive
  & joint 
  & Waymo  
  & monolithic decoder-only transformer with unified token space for motion and map\\ 

UniMM \cite{lin_revisit_2025} 
  & 2025 
  & Autoregressive
  & joint 
  & Waymo 
  & decoder-only transformer family for continuous or hybrid Gaussian mixture prediction\\

CTG \cite{zhong_guided_2023} 
  & 2023 
  & Diffusion 
  & marginal 
  & nuScenes 
  & controlled diffusion with agent-centric scene encoding\\

MotionDiffuser \cite{jiang_motiondiffuser_2023}  
  & 2023 
  & Diffusion 
  & joint  
  & Waymo 
  & joint controlled diffusion with agent-centric scene encoding\\
  
DICE \cite{choi_dice_2024}  
  & 2024 
  & Diffusion 
  & marginal 
  & ETH-UCY, nuScenes 
  & diffusion with plausibility scoring and selection\\

VBD \cite{huang_versatile_2024} 
  & 2024 
  & Diffusion 
  & marginal/joint 
  & Waymo 
  & behavior prior predictor and controlled diffusion with query-centric scene encoding\\

\bottomrule
\end{tabular}}}
\end{table*}

\paragraph{Autoregressive Models}
Autoregressive modeling approaches factorize the distribution over motion sequences as a product of conditionals.
This factorization preserves the natural temporal causality found in motion sequences and allows to model the interaction between agents and environmental elements (e.g. traffic lights) at every step of the conditional roll-out.
Early pioneering works, such as MFP \cite{tang_multiple_2019} adopted this approach with \acrshortpl{rnn}.
However, their adaptation slowed due to training instabilities.
The necessarily sequential roll-outs lead to cumulative errors, making long-term dependencies brittle and resulting in high sampling latencies.

The success of \acrshortpl{gt} in the \acrfull{nlp} domain sparked a renewed interest.
Drawing inspiration from sequence-to-sequence architectures in the \acrshort{nlp} domain, multiple autoregressive Transformer encoder-decoder architectures have been proposed, including LatentFormer \cite{abolfathi_latentformer_2022}, 
MotionLM \cite{seff_motionlm_2023} and AMP \cite{jia_amp_2024}.
These models combine domain-specific encoding strategies, e.g. from \acrfull{sota} discriminative models such as Wayformer \cite{nayakanti_wayformer_2023} or MTR \cite{chen_motiontransformer_2019}, with token-based decoding strategies from the \acrshort{nlp} domain.
Tokenization enables a unified representation of inputs and outputs.
Advances from \acrshort{nlp}, such as attention mechanisms, training objectives, and sampling techniques, have directly improved training stability and reduced sampling latency.
Trajeglish \cite{philion_trajeglish_2024} discretizes trajectories into GPT-like small vocabulary token set, which are then autoregressively predicted using discrete sequence modeling methods. It depicts that language-like encoding can be extended for application in trajectory forecasting domain. Apart from using language-like tokenizers, motion tokens as words in natural language can be used to make trajectory forecasting model's hidden states interpretable, and to control the motion forecasts \cite{tas_words_2025}.
Finally, \cite{lin_revisit_2025} proposes a unified framework based on mixture models to predict or generate traffic behavior, including continuous models and discrete GPT-like models as special cases.
\cite{wu_smart_2024} trains a causal-decoder transformer on motion tokens from multiple datasets, demonstrating effective inference without loss of performance.

\paragraph{\acrfullpl{dm}}
Recent research has explored the capabilities of \acrshortpl{dm} in trajectory forecasting and generation, highlighting their versatility and effectiveness. 

MotionDiffuser \cite{jiang_motiondiffuser_2023} and DICE \cite{choi_dice_2024} add positional noise to future trajectories. They train an encoder-denoiser architecture where the encoder conditions on context and the decoder (denoiser) takes the conditioned information, predicts and removes the noise to recover the trajectory. In another work, \cite{zhong_guided_2023}  models trajectories as action-state sequences and employs a \acrshort{dm} to denoise actions, thereby reconstructing the trajectory. The method underlines the flexibility of \acrshortpl{dm} across representations. 
The works \cite{jiang_motiondiffuser_2023} and \cite{zhong_guided_2023} outline another benefit of \acrshortpl{dm} with respect to control of the trajectory generation process. Besides conditioning on the scene context to produce scene-consistent trajectories, these works use an online gradient-based guidance mechanism during the diffusion process for controllable trajectory generation. Originating from Diffuser \cite{janner_planning_2022}, the key idea is to use the gradient method during the denoising process to meet high-level specifications for the generated trajectories. While \cite{zhong_guided_2023} introduces guidance to comply with the safety and traffic rules specified by \acrshort{stl} formulas \cite{maler_monitoring_2004}, the work in \cite{jiang_motiondiffuser_2023} provides a framework for performing controlled trajectory sampling based on attractor or repeller functions in the target position. Another work in \cite{huang_versatile_2024} develops a method to enhance the diversity of traffic scenarios for \acrshort{ad}, proposing a pursuit-evasion game in the guidance policy to generate safety-critical multi-agent trajectories.

These advances illustrate the potential of \acrshortpl{dm} in improving the accuracy and reliability of trajectory forecasting and generation, making them invaluable tools in fields that require precise trajectory forecasting and environmental understanding. However, we find that \acrshort{fm} remain largely unexplored for AD trajectory forecasting.

\paragraph{\acrfullpl{ebm}} \acrshortpl{ebm} have been successfully applied to pedestrian motion prediction. \cite{pang_trajectory_2021} introduces an approach for predicting human trajectories using an \acrshort{ebm}. The authors present a latent belief mechanism to capture uncertainty and interactions among agents. In \cite{wang_seem_2022}, the Sequence Entropy Energy-based Model (SEEM) is proposed as a model composed of a \acrfull{lstm}-based generator network and an energy network to determine the most representative trajectory. The more recent paper \cite{liu_modeling_2024} addresses the challenge of predicting pedestrian trajectories in uncertain environments taking a hybrid approach combining \acrshortpl{dm} and \acrshortpl{ebm}. The method includes the use of \acrshortpl{ebm} to represent potential future trajectories and a denoising strategy that refines these predictions to more plausible ones.

\subsection{Trajectory Forecasting Metrics}\label{subsec:pred_metrics}
Forecasting models need to be assessed on multiple dimensions (cf. Table~\ref{tab:traj-pred-metrics}). Most important are displacement metrics, measuring the lateral and longitudinal alignment of predictions to given ground truths.  Suitable metrics are average- or final displacement error (ADE)/(FDE) and their oracle forms (minADE/minFDE@$K$), measuring the quality of a single trajectory or the best out of $K$. MissRate@$K,\delta$, measures the fraction of cases where no top-$K$ prediction lies within $\delta$ m of the ground truth. \cite{huang_survey_2022}

However, due to the inherent multi-modal nature of trajectory forecasting, it is also important to evaluate the distribution of predicted trajectories. Probabilistic fit is quantified with NLL \cite{huang_survey_2022}, or Brier variants of FDE/ADE. Expected Calibration Error (ECE) \cite{pakdaman_naeini_obtaining_2015} evaluates the discrepancy between the predicted confidence of a trajectory and the probability of correctness. Diversity of modes is multi-modal forecast settings is measured through top-$K$ coverage or pairwise spread, measuring the distance between modes \cite{shiroshita_behaviorally_2020}. 

Additionally, some compliance measures exist, measuring alignment to criteria based on regulatory or social norms. Drivable area compliance \cite{chang_argoverse_2019} is a common example, but few papers also introduce more scenario specific variants. Composite \acrshort{nll} \cite{montali_waymo_2023} aggregates multiple compliance measures by measuring the alignment between predictions and ground truth compliance to achieve an empirical measure of realism.

\begin{table}[t]
\centering
\caption{Trajectory Forecasting Metrics}
\label{tab:traj-pred-metrics}
\small
\resizebox{\columnwidth}{!}{%
\begin{tabular}{@{}lll@{}}
\toprule
\textbf{Category} & \textbf{Metric} & \textbf{Description} \\
\midrule
\textbf{Displacement} & minADE@$K$ \cite{huang_survey_2022}
& Average Displacement  \\

& minFDE@$K$ \cite{huang_survey_2022}
& Displacement of final step \\

& MissRate@$\delta$ \cite{huang_survey_2022} 
& Existence of prediction within bound \\
\midrule
\textbf{Probabilistic} & \acrfull{nll} \cite{huang_survey_2022} & 
Probability of ground truth \\

& Expected Calibration Error \cite{pakdaman_naeini_obtaining_2015}
& Reliability of predicted confidences \\
\midrule
\textbf{Multi-Modality} & Pairwise-Spread \cite{shiroshita_behaviorally_2020}
& Spread of predicted modes  \\
& Top-K Coverage \cite{shiroshita_behaviorally_2020} &  \\
\midrule
\textbf{Compliance} & Drivable Area Compliance \cite{chang_argoverse_2019}
& Amount of way points not on roads \\
\midrule
\textbf{Composite} & Composite \acrshort{nll} \cite{montali_waymo_2023} & Aggregation of probabilistic metrics \\
\bottomrule
\end{tabular}}
\end{table}

\section{Motion Planning}
\label{sec:planning}
\begin{table*}[h!]
    \centering
    \caption{Overview of Hybrid Motion Planning Methods}
    \resizebox{0.9\textwidth}{!}{%
    \begin{tabular}{@{}lcccc@{}}
        \toprule
        \textbf{Method} & \textbf{Year} & \textbf{Approach} & \textbf{Benchmarks} & \textbf{Features} \\
        \midrule
        Performer-MPC~\cite{xiao_learning_2022} & 2022 & Transformer + MPC & Custom Sim/Real World & Learns cost function from vision embeddings; robust to constraint violations. \\
        GAN-MPC~\cite{burnwal_gan-mpc_2023} & 2023 & GAN + MPC &  DeepMind Control suite & Minimizes the Jensen-Shannon divergence between the state-trajectory distributions. \\
        LanguageMPC~\cite{sha_languagempc_2023} & 2023 & LLM + MPC & SUMO + IDSim & LLM generates MPC parameters and cost weights from text prompts. \\
        Transformer-MPC~\cite{lotfi_uncertainty-aware_2024} & 2024 & Transformer + MPC & Real World & Replaces BADGR’s LSTM with Transformer; improves uncertainty-aware control. \\
        Diffusion SDE MPC~\cite{djeumou_one_2025} & 2025 & Diffusion + MPC & Real World & Physics-informed conditional diffusion for stochastic vehicle dynamics. \\
        D-MPC~\cite{zhou_diffusion_2024} & 2024 & Diffusion + MPC & D4RL & Generates initial trajectory; MPC refines under dynamic constraints. \\
        CGAN-RRT*~\cite{ma_conditional_2022} & 2022 & CGAN + RRT* & Custom Sim & CGAN guides RRT* sampling; accelerates convergence. \\
        NF-A2DF/-AIL~\cite{rabenstein_sampling_2024} & 2024 & NF + MPPI & CarMaker & Learns control input distribution for importance sampling. \\
        Co-Pilot~\cite{wang_chatgpt_2023} & 2023 & LLM + MPC/PID & Simulink + CarSim & ChatGPT is applied to two different tasks: path tracking control and trajectory planning. \\
        CoBL-Diffusion~\cite{mizuta_cobl-diffusion_2024} & 2024 & Diffusion + CBF/CLF & Custom Sim & Integrates CBF and CLF for safety and stability within the diffusion process. \\
        SafeDiffuser~\cite{xiao_safediffuser_2023} & 2024 & Diffusion + CBF & MuJoCO/Pybullet & Applies safety filter during denoising for collision avoidance. \\
        DriveVLM-Dual~\cite{tian_drivevlm_2024} & 2024 & VLM + Planner & nuScenes / Real World & Chain-of-thought scene reasoning; hierarchical planning refinement. \\
        LLM-A*~\cite{meng_llm-_2024} & 2025 & LLM + A* & Custom Sim & Using LLMs to reduce computational and memory limitations. \\
        \bottomrule
    \end{tabular}%
    }
    \label{tab:mot_plan_overview}
\end{table*}

The goal of motion planning is to determine appropriate behavior for an AD system considering the current state of the vehicle and the environment and key criteria such as safety, feasibility and comfort. In this section, we discuss classical motion planning methods and examine how generative models can enhance them, while \Cref{sec:end-to-end} covers \acrshort{e2e} architectures.

\subsection{Overcoming Conventional Planner Challenges}
\label{subsec:overcoming-conventional-planner-challenges}

In recent years, classical planning methods have been widely adopted and can be broadly divided into optimization-based and search- or sampling-based approaches. The former is dominated by \acrfull{mpc} techniques \cite{micheli_nmpc_2023, tas_decision-theoretic_2024}, while the latter includes lattice- and A*-based planners, Monte Carlo Tree Search, and RRT-based methods \cite{somani_despot_2013, boroujeni_flexible_2017, qi_mod-rrt_2021}.
Although these methods usually provide strict constraint satisfaction, their performance is often limited by high computational cost. As a result, they rely on simplifying assumptions, short planning horizons, or compromised  performance \cite{bouzidi_learning-aided_2024}. Moreover, search- and sampling-based methods typically use coarse discretizations in time and action spaces, which can yield sub-optimal solutions in highly dynamic environments~\cite{chen_interactive_2022, bouzidi_motion_2024}.
One common modeling simplification is to ignore the interactions in traffic by decoupling prediction from planning \cite{trautman_unfreezing_2010, bouzidi_interaction-aware_2023}. This approach can cause \acrshort{ad} systems to behave in a reactive manner or to freeze altogether in scenarios with high uncertainty. On the other hand, learning-based methods excel due to their ability to account for the complex and uncertain nature of other agents.

Common learning-based methods \cite{tordesillas_deep-panther_2023, nishi_merging_2019} leveraging \acrshort{rl} or \acrshort{il} are fast during inference and can adapt to complex environments, but they need an exhaustive exploration of the state space during learning to perform reliably. As a result, these methods struggle with the "long-tail" scenarios outside their training data, where a lack of generalization can undermine reliability. The lack of generalization capabilities, safety guarantees and interpretability is a concern in such safety-critical applications and is further discussed in \Cref{subsec:challenges}. Furthermore, integrating human priors is challenging, as human knowledge and reasoning are expressed in language-based representations, whereas these systems primarily process numerical sensory data \cite{fu_drive_2023, mao_gpt-driver_2023}.  

Generative models offer promising solutions to these limitations by providing a deeper contextual understanding and robust generalization capabilities \cite{hallgarten_can_2024}. These models can incorporate multimodal knowledge, allowing them to interpret complex, dynamic environments more holistically and to generate "what-if" scenarios that support informed decision-making in uncertain conditions. Unlike conventional systems that are limited to numeric inputs, generative models can represent human-like priors and reasoning, enabling planning and decisions that resemble human intuition and foresight \cite{huang_making_2024, hallgarten_can_2024,mao_gpt-driver_2023}. This capability allows generative models to learn complex, multimodal behavior distributions of multiple agents, making them well-suited for interactive scenarios.

\subsection{Hybrid Methods}\label{subsec:hybrid-methods}
Due to the complementary features of classical and learning-based planning methods (\Cref{subsec:overcoming-conventional-planner-challenges}), many hybrid approaches have been proposed. See Table~\ref{tab:mot_plan_overview} for an overview of representative methods. These methods generally fall into two categories. The first uses learning to replace or enhance components of classical planners that are difficult to tune manually. Some examples include learning cost function weights and shaping the cost \cite{zarrouki_weights-varying_2021, brito_where_2021}, or learning system dynamics to handle complex behavior \cite{hewing_learning-based_2020}. The second category learns high-level policies that are refined by low-level model-based planners such as \acrshort{mpc}. This includes policy-guided \acrshort{mpc} \cite{brito_learning_2022, song_policy_2021}, safety filters \cite{wabersich_predictive_2023}, and learning-based warm-starting of \acrshort{mpc} \cite{natarajan_learning_2021, bouzidi_learning-aided_2024}. Consequently, recent works explore combinations of \acrshort{genai} and classical approaches.

\paragraph{ Generative Methods for Planner Subcomponents}
Performer-\acrshort{mpc} \cite{xiao_learning_2022} uses a learned \acrshort{mpc} 
cost function parametrized by vision context embeddings provided by Performers, a low-rank implicit-attention Transformer. This work combines the benefits of \acrshort{il} with robust handling of system constraints in \acrshort{mpc}. 
Similarly, GAN-MPC \cite{burnwal_gan-mpc_2023} employs a GAN to minimize the Jensen-Shannon divergence between the state-trajectory distributions of the demonstrator and the imitator. This method does not require the demonstrator and the imitator to share the same dynamics, and their state spaces may have a partial overlap. 
In \cite{sha_languagempc_2023} the authors use an \acrshort{llm} to generate weight matrices, observation matrices, and action biases for \acrshort{mpc} based on textual prompts. The weights shape the cost function of the \acrshort{mpc}, while the observation matrix selects the vehicles considered as obstacles. The action bias corresponds to the action directive of the \acrshort{llm} (e.g., stop).

The authors in \cite{lotfi_uncertainty-aware_2024} substituted the \acrshort{lstm} model within an \acrshort{mpc} from the BADGR method (Berkeley Autonomous Driving Ground Robot) with a transformer model, resulting in significant performance improvement. Recent work \cite{djeumou_one_2025} proposed a diffusion-based approach to learning a physics-informed state-space model for \acrshort{mpc}. The method employs a conditional \acrshort{dm} to capture the multimodal distribution of vehicle dynamics, enabling high-performance driving under uncertainty, thereby enhancing safety. By conditioning on current state measurements, the \acrshort{dm} outputs the parameters of a physics-informed neural \acrshort{sde}, which is used as the state-space model within the \acrshort{mpc}.

A \acrshort{dm} is used in \cite{zhou_diffusion_2024} to improve \acrshort{mpc} performance by learning both the dynamic model (or world model) of the \acrshort{mpc} and an initial guess for the optimization procedure, tackling two major challenges in \acrshort{mpc}. The trajectory generated by the \acrshort{dm} planner is thus refined by the \acrshort{mpc}, considering the respective constraints to output a feasible trajectory. 

Some recent works have explored the synergy of \acrshort{genai} and search- or sampling-based methods. CGAN-RRT* \cite{ma_conditional_2022} combines \acrfullpl{cgan} with the RRT* algorithm. The CGAN model generates a distribution of feasible paths conditioned on the input map, which guides the sampling process of RRT* to accelerate convergence to an optimal path. Similarly, \cite{rabenstein_sampling_2024} leverages \acrshortpl{nf} to learn a distribution of control inputs used by the \acrfull{mppi} planner to find the optimal rollout via importance sampling.

\paragraph{ Generative Methods for Trajectory Proposals}

Focusing on highway driving, Wang et al. \cite{wang_empowering_2024} combine \acrshort{mpc} with an \acrshort{llm}. The OpenAI GPT-4 model is used as the driver agent and takes as input the environment state as a textual description. The LLM outputs the desired lane, serving as input for the low-level \acrshort{mpc}. This is further extended for interactive planning by integrating a state machine, which provides a rule-based framework for guiding the LLM and the prediction of the surrounding vehicles, resulting in enhanced safety and interpretability.

CoBL-Diffusion \cite{mizuta_cobl-diffusion_2024} and SafeDiffuser \cite{xiao_safediffuser_2023} utilize the concept of \acrfullpl{cbf} as a safety filter for diffusion-based planners. They refine the trajectories through a guided diffusion process while still enforcing safety constraints through the \acrshortpl{cbf}, integrating it into the denoising process, which helps predict dynamically feasible paths by, e.g., minimizing collision risks with moving obstacles. Similarly, DDM-Lag \cite{liu_ddm-lag_2024} adopts a Lagrangian-based constrained optimization step to integrate safe collision avoidance into the reinforcement-guided diffusion policy learning.

A two-stage hybrid planner is also introduced in \cite{hallgarten_can_2024} that combines an LLM with a rule-based planner. In the first stage, the \acrshort{llm} planner processes a natural language prompt containing task instructions, perception context, ego states, and route information to select a suitable driving behavior, such as lane following, or overtaking obstacles. The selected behavior is then passed to the \acrfull{pdm}. The \acrshort{pdm} is responsible for refining the high-level maneuver by sampling trajectories and selecting the best one based on a minimum cost. The work indicates that this combination can outperform purely LLM-based and purely rule-based planners.

DriveVLM \cite{tian_drivevlm_2024} enhances trajectory prediction via chain-of-thought reasoning, combining scene description, analysis, and hierarchical planning. Detected objects’ 3D bounding boxes are converted into language prompts to boost spatial understanding. A two-stage hybrid planner refines low-frequency generative trajectories in real-time using a traditional planner.

LLM-A* \cite{meng_llm-_2024} combines an LLM with the traditional A* algorithm. The LLM provides high-level sub-goals for guiding the search based on the start state, goal state, and obstacles. The A* planner then plans a trajectory inbetween these subgoals, ensuring optimality and validity. Conversely, the generative model reduces the large computational cost of A* by decreasing the number of states explored.

\subsection{Motion Planning Metrics}

Motion planning performance is typically evaluated in closed-loop settings, such as simulations or real-world tests, to measure its effect on vehicle behavior. We categorize common metrics in \Cref{tab:motion-metrics} into safety, compliance, efficiency and comfort \cite{alban_getting_2025, karnchanachari_towards_2024}. Safety and compliance metrics are typically considered primary, whereas efficiency and comfort metrics provide insight into the overall quality of motion. Benchmarks such as nuPlan \cite{dauner_parting_2023} and CARLA \cite{dosovitskiy_carla_2017} combine these indicators into a weighted score, where violations of safety constraints generally result in a zero scenario score. While learning-based planners have popularized open-loop metrics akin to those used in motion prediction presented in \Cref{subsec:pred_metrics}, recent studies \cite{dauner_parting_2023, hallgarten_can_2024} highlight their limitations in reflecting true closed-loop driving performance.

\begin{table}[t]
\centering
\caption{Overview of Motion Planning Metrics}
\small
\resizebox{\columnwidth}{!}{%
\begin{tabular}{@{}llp{7cm}@{}}
\toprule
\textbf{Category} & \textbf{Metric} & \textbf{Description} \\
\midrule

\textbf{Safety} 
& Collision rate \cite{karnchanachari_towards_2024}
& Fraction of scenarios with any collision. \\

& Time-to-collision (TTC) \cite{karnchanachari_towards_2024}
& Minimum predicted time to impact. \\

& Drivable area compliance \cite{karnchanachari_towards_2024}
& Fraction of trajectory within drivable space. \\

& Safe following distance (SFD) \cite{alban_getting_2025}
& Safe gap behind vehicles to react and stop safely. \\

\midrule
\textbf{Compliance} 
& Lane deviation \cite{alban_getting_2025}
& Lateral deviation from the assigned lane center. \\

& Speed-limit compliance \cite{karnchanachari_towards_2024}
& Fraction of time that ego remains within speed limits. \\

& Traffic-light / sign violation \cite{alban_getting_2025}
& Binary or count-based indicator of rule violations. \\

& Right-of-way compliance \cite{alban_getting_2025}
& Whether behavior follows intersection priority rules. \\

\midrule
\textbf{Efficiency} 
& Path progress \cite{alban_getting_2025}
& Distance advanced along the global route. \\

& Time-to-goal \cite{alban_getting_2025}
& Time required to reach the scenario goal. \\

\midrule
\textbf{Comfort} 
& Acceleration / deceleration \cite{alban_getting_2025}
& Mean and maximum longitudinal/lateral acceleration. \\

& Jerk \cite{alban_getting_2025}
& Time derivative of acceleration. \\

\bottomrule
\end{tabular}\label{tab:motion-metrics}
}
\end{table}

\section{End-to-End Driving}
\label{sec:end-to-end}
This section explores how \acrshort{e2e} models address limitations of traditional modular motion planning in \acrshort{ad}. We categorize recent generative \acrshort{e2e} approaches by \acrshortpl{dm}, \acrshortpl{vae}, Transformers and \acrshortpl{llm}, and discuss key challenges and open issues in real-world deployment.

\begin{table*}[h!]
    \centering
    \caption{Overview of End-to-End Methods}
    \resizebox{0.9\textwidth}{!}{%
    \begin{tabular}{@{}lccccc@{}}
        \toprule
        \textbf{Model} & \textbf{Year}  & \textbf{Approach} & \textbf{OL / CL} & \textbf{Benchmarks} & \textbf{Features} \\
        \midrule
        DiffusionDrive \cite{liao_diffusiondrive_2024} & 2025 & Diffusion & OL & nuScenes, NAVSIM & Diffusion with anchored Gaussian priors and transformer-based decoder for conditioning.\\
        GenAD \cite{leonardis_genad_2025} & 2024 & Transport, Autoregressive  & OL & nuScenes & Encodes scene features into a latent trajectory space, then predicts\&plans future trajectories from it. \\
        DriveGPT4 \cite{xu_drivegpt4_2024} & 2024 &  Autoregressive & OL & BDD-X & Predicts vehicle control and driving reasoning.\\
        OmniDrive \cite{wang_omnidrive_2024} & 2025 & Autoregressive & OL & nuScenes & Leverages 2D VLM priors with 3D position encoding and counterfactual reasoning.\\
        SimLingo \cite{renz_simlingo_2025} & 2025 & Autoregressive & CL & Carla, Bench2Drive & Maps split front-view image into path and waypoint representations.\\
        DriveLM \cite{tian_drivevlm_2024} & 2024 &  Autoregressive & OL & nuScenes, SUP-AD &Performs scene-level reasoning for improved planning.\\
        EMMA \cite{hwang_emma_2024} & 2024 &  Autoregressive & OL & nuScenes, internal dataset &  Applies chain-of-thought prompting using visual and historical context.\\
        Senna \cite{jiang_senna_2024} & 2024 &  Autoregressive & OL & nuScenes, DriveX &Uses a \acrshort{vlm} for high-level planning with a separate low-level prediction module.\\
        BEVDriver \cite{winter_bevdriver_2025} & 2025 &  Autoregressive & OL + CL & LangAuto (Carla) & Feeds \acrshort{bev} features and navigation instructions to an LLM to generate future waypoints.\\
        \bottomrule
    \end{tabular}}
    \label{tab:e2e-models}
\end{table*}

Conventional \acrshort{ad} architectures consist of modular pipelines with task-specific components for perception, prediction, and planning (cf. \Cref{fig:ad_stack}), where outputs feed sequentially into subsequent modules \cite{levinson_towards_2011, behere_functional_2016, munir_autonomous_2018}. These systems are interpretable, easier to verify, and allow for failure tracing and module adjustments \cite{tas_automated_2017, tampuu_survey_2022}. However, modular designs are hindered by abstract, hand-crafted interfaces, limiting their ability to generalize across diverse scenarios and producing outputs like 3D bounding boxes that may be insufficient for decision-making in complex \acrshort{ad} scenarios \cite{tampuu_survey_2022, dosovitskiy_carla_2017, jiang_senna_2024}. Additionally, error propagation from upstream modules, such as missed objects in detection, compromises downstream planning \cite{dosovitskiy_carla_2017}. Lastly, these architectures struggle to account for interactions between the ego-vehicle and other agents, as the downstream propagation of information prevents direct integration of ego motion in prediction.

\acrshort{e2e} \acrshort{ad} refers to a streamlined motion planning approach using a \acrfull{dnn} to process sensor data and output low-level control signals like steering and acceleration. The model directly learns the driving task, with intermediate feature representations trained jointly, eliminating the need for complex module interfaces \cite{chib_recent_2024, chen_end--end_2024, tampuu_survey_2022, jiang_senna_2024}. While most \acrshort{e2e} systems use a single large network, they can also include submodules optimized jointly for the driving task \cite{chen_end--end_2024}. Related tasks like detection, tracking, prediction, and planning are often addressed simultaneously \cite{hu_planning-oriented_2023, sun_sparsedrive_2024}.

\acrshort{e2e} models are considered less complex than modular pipelines, as they feature fewer intermediate representations and a streamlined architecture. These models are more robust in handling diverse scenarios, learning implicit patterns that predefined module chains may overlook. 
Additionally, \acrshort{e2e} models are computationally efficient, benefiting from the elimination of redundant computations and the use of a shared backbone \cite{chib_recent_2024, chen_end--end_2024}. Unlike rule-based approaches, these data-driven models scale effectively with data and training resources, enhancing their adaptability and performance \cite{chen_end--end_2024}.

\subsection{Current Methods}

State-of-the-art \acrshort{e2e} architectures for \acrshort{ad} typically employ generative \acrshort{dnn} backbones, often vision-centric or incorporating Transformers for attention-based spatial and temporal reasoning. Table~\ref{tab:e2e-models} provides an overview of representative methods in this field.

\paragraph{\acrfullpl{dm}}
DiffusionDrive \cite{liao_diffusiondrive_2024} adopts a truncated diffusion policy that generates a multi-mode action distribution from a prior anchored Gaussian distribution of trajectories, streamlining the denoising process and enhancing real-time performance and a transformer-based diffusion decoder to improve scene context conditioning.

\paragraph{\acrshortpl{vae}}
GenAD \cite{leonardis_genad_2025} uses self- and cross-attention to generate an instance-centric \acrshort{bev} scene representation, which is mapped with latent trajectory priors learned with a \acrshort{vae} to step-wise predict the next future with gate recurrent units and decoding them into explicit trajectories.

\paragraph{LLM-based Models}
The advanced reasoning skills and contextual understanding of \acrshortpl{llm} suit these models for decision making and planning units.
Furthermore, by leveraging their natural language understanding capabilities, \acrshortpl{llm} can offer explainability, addressing the broader challenge of interpretability in \acrshort{e2e} models. However, it is an ongoing research problem to improve their spatial understanding and real-time feasibility.
DriveGPT4 \cite{xu_drivegpt4_2024} adopts LLama-2 for \acrshort{e2e} planning, demonstrating high predictive performance and interpretability through action descriptions and justifications using natural language. 
OmniDrive \cite{wang_omnidrive_2024} utilizes a 3D position encoding for perception tasks leveraging 2D pretrained knowledge combined with counterfactual reasoning.  
SimLingo \cite{renz_simlingo_2025} leverages a \acrshort{vlm} optimized for closed-loop driving, integrating spatially aligned features from two front-view image splits into a TinyLLaMA decoder with 50M parameters. It efficiently generates paths and waypoints in a disentangled representation for improved control, demonstrating outstanding performance. DriveLM \cite{sima_drivelm_2024} uses Graph Visual Question Answering, a technique that mimics a human decision making process in driving by linking different reasoning steps, such as perception, prediction, planning, behavior and motion, into a graph-based logical chain. 
EMMA \cite{hwang_emma_2024} yields competitive performance in the NuScenes and Waymo benchmarks by using Gemini's \cite{team_gemini_2024} world knowledge to produce Chain-of-Thought reasoning and future waypoints.
Senna \cite{jiang_senna_2024} combines a \acrshort{vlm} with an \acrshort{e2e} framework based on VADv2 \cite{chen_vadv2_2024}, which generates meta-actions using reasoning and driving knowledge to guide the \acrshort{e2e} model in producing precise trajectory predictions. BEVDriver \cite{winter_bevdriver_2025} feeds latent \acrshort{bev} maps from LiDAR and multi-view images as an efficient spatial representation to an LLM, which generates waypoints that incorporate high level decisions.

\acrshort{e2e} driving models are evaluated on the same metrics as hybrid and modular motion planning methods (see~\Cref{tab:motion-metrics}).

\section{Datasets and Simulators}
\label{sec:data}
This section overviews datasets for trajectory forecasting and motion planning. We introduce some real-world datasets and discuss commonly used driving simulators that can be used for research in \acrshort{ad}.%to record synthetic datasets or for experiments in \acrshort{rl}.
\subsection{Datasets usable for above categories}
\begin{table*}[h!]
    \centering
    \caption{Overview of Motion Datasets \cite{wilson_argoverse_2023}}
       \resizebox{0.9\textwidth}{!}{\begin{tabular}{@{}lccccccc@{}}
        \toprule
        & \textbf{Argoverse}~\cite{chang_argoverse_2019}  & \textbf{INTER.}~\cite{zhan_interaction_2019}& \textbf{LYFT}~\cite{houston_one_2021}& \textbf{Waymo}~\cite{ettinger_large_2021} &\textbf{NuScenes}~\cite{caesar_nuscenes_2020}& \textbf{Yandex}~\cite{malinin_shifts_2022} & \textbf{Argoverse2}~\cite{wilson_argoverse_2023} \\ \midrule
        \# Scenarios & $324$ k & $-$ & $170$ k & $104$ k & $41$ k & $600$ k & $250$ k \\
        \# Unique Tracks & $11.7$ M & $40$ k & $53.4$ M & $7.64$ M & $-$ & $17.4$ M & $13.9$ M \\
        Average Track Length & $2.48$ s & $19.8$ s & $1.84$ s & $7.04$ s & $-$ & $-$ & $5.16$ s \\
        Total Time & $320$ h & $16.5$ h & $1118$ h & $574$ h & $5.5$ h & $1667$ h & $763$ h \\
        Scenario Duration & $5$ s & $-$ & $25$ s & $9.1$ s & $8$ s & $10$ s & $11$ s \\
        Test Forecast Horizon & $3$ s & $3$ s & $5$ s  & $8$ s & $6$ s & $5$ s & $6$ s \\
        Sampling Rate & $10$ Hz & $10$ Hz & $10$ Hz & $10$ Hz & $2$ Hz & $5$ Hz & $10$ Hz \\
        \# Cities & $2$ & $6$ & $1$ & $6$ & $1$ & $6$ & $6$ \\
        Unique Roadways & $290$ km & $2$ km & $10$ km & $1750$ km & $-$ & $-$ & $2220$ km \\
        Avg. \# tracks per scenario & $50$ & $-$ & $79$ & $-$ & $75$ & $29$ & $73$ \\
        \# Evaluated object categories & $1$ & $1$ & $3$ & $3$ & $1$ & $2$ & $5$ \\
        Multi-agent evaluation & $\times$ & $\checkmark$ & $\checkmark$ & $\checkmark$ & $\times$ & $\checkmark$ & $\checkmark$ \\
        Mined for interestingness & $\checkmark$ & $\times$ & $-$ & $\checkmark$ & $\times$ & $\times$ & $\checkmark$ \\
        Vector Map & $\checkmark$ & $\times$ & $\times$ & $\checkmark$ & $\checkmark$ & $\times$ & $\checkmark$ \\
        Download Size & $4.8$ GB & $-$ & $22$ GB & $1.4$ TB & $48$ GB & $120$ GB & $32$ GB \\
        \# Public Leaderboard Entries & $194$ & $-$ & $935$ & $23$ & $18$ & $3$ & $-$ \\ \bottomrule
    \end{tabular}}
    \label{tab:comparison}
\end{table*} Datasets with agent motion, often connected to sensor data, are foundational for \acrshort{genai}. On the one hand, they are used to train GenAI models directly. On the other hand, they serve pre-training purposes and enable the extraction of meaningful representations of street scenes. \Cref{tab:comparison} presents an overview of different motion datasets usable for scenario generation, trajectory forecasting and motion planning models. Additionally, it compares different properties, such as the number of scenarios and the total recording time. The datasets are recorded with various sensors, including RGB camera, radar, and LiDAR. Furthermore, they are preprocessed by reducing the resolution \cite{wilson_argoverse_2023} or deleting recordings with onboard system failures \cite{malinin_shifts_2022}. Argoverse~\cite{chang_argoverse_2019} is a dataset for trajectory forecasting that was recorded using LiDAR, RGB video sensors, front-facing stereo, and a 6-DOF localization. It includes data from American cities in different seasons, weather conditions, and times of day. Argoverse is preferred for 3D tracking and trajectory forecasting. 
In INTERACTION~\cite{zhan_interaction_2019}, drone and traffic cameras were used to record scenarios.
A big advantage is that INTERACTION is critical and complex, e.g.\ unstructured and chaotic scenarios, which lead to dangerous maneuvers.
It has diverse scenarios like highway, roundabout, stopping at traffic signs, and lane changes. LYFT~\cite{houston_one_2021} provides data from cameras, LiDAR, and radar sensors, as well as semantic \acrshort{hd}-Maps for trajectory forecasting. Those \acrshort{hd}-Maps capture road rules, lane geometry, and other traffic elements. However, LYFT only contains daytime data. Similarly, Waymo~\cite{ettinger_large_2021} is a large-scale dataset for forecasting the trajectories of other vehicles by providing 3D bounding boxes and map data in diverse scenes. The NuScenes~\cite{caesar_nuscenes_2020} dataset was recorded using cameras, radar, and LiDAR with a $360^{\circ}$ field of view. Notably, radar sensor data is rarely included in other datasets. Additionally, NuScenes provides diverse situations like comparing left and right-hand driving. 
Likewise, the Yandex~\cite{malinin_shifts_2022} dataset is released for trajectory forecasting and is collected using camera-based sensors. To ensure high data quality, the raw recordings are preprocessed to remove scenes exhibiting onboard system failures or violating physical constraints. Additionally, it contains a wide range of scenes, including different countries (Israel, United States, and Russia), with different seasons, times of day, and weather conditions. Yandex is also released for evaluating uncertainty estimation and robustness to domain gaps. Therefore, it provides an in-domain dataset and an out-domain dataset.
Finally, Argoverse2~\cite{wilson_argoverse_2023} is a dataset for trajectory forecasting. Besides the forecasting motion dataset, it contains a sensor dataset with LiDAR data. It is recorded in diverse cities of the United States with new challenging traffic situations, different types of vehicles and more diverse weather conditions compared to Argoverse~\cite{chang_argoverse_2019}. 

\subsection{Driving simulators }
Driving simulators are a valuable option for experiments in \acrshort{ad} to explore and analyze dangerous situations, e.g. collision scenarios with pedestrians. 
Furthermore, they support closed-loop model training. Finally, synthetic data are cheaper than real data.
CARLA~\cite{dosovitskiy_carla_2017} is an open-source \acrshort{ad} simulator. It is based on Unreal Engine~\cite{noauthor_unreal_2014} and its roads and urban settings are defined by using the ASAM OpenDrive standard~\cite{bassermann_asam_2025}. Because CARLA is developed for \acrshort{ad} research, it provides different sensors including cameras, LIDAR sensors, and provides semantic and instance segmentation. 
For \acrshort{ad} tasks, CARLA is used in \acrshort{rl} \cite{li_think2drive_2024,
%gutierrez-moreno_reinforcement_2022,
hossain_autonomous_2023}, trajectory forecasting~\cite{sun_p4p_2022} and motion planning~\cite{wu_trajectory-guided_2022,roh_multimodal_2020}.

SUMO~\cite{lopez_microscopic_2018} is an open-source driving simulator focused on traffic scenario generation and research~\cite{markowski_interfacing_2024}, as well as vehicle-to-vehicle communication~\cite{naeem_vehicle_2020}. It enables the analysis of different complex road networks and vehicles. The road network can be constructed by the user, who can build their own network or use real road networks, which can be included by e.g.\ OSM~\cite{planetosm_planet_2012} or XML-Descriptions. SUMO only provides a \acrshort{bev} on a 2D-world.
However, pedestrians and information about the vehicles i.e.\ emissions are available.   
SUMO is also used for \acrshort{rl} tasks~\cite{xu_deep_2024,
%zhang_cityflow_2019,
makantasis_deep_2020} whereby the focus is rather on the traffic scenario itself i.e.\ traffic light controlling than on one ego-vehicle. Furthermore, it is applied for trajectory forecasting~\cite{klischat_coupling_nodate,artunedo_motion_2020} and motion planning~\cite{gu_trajectory_2024}.

TORCS~\cite{espie_torcs_2005} is a racing simulator that provides a diverse range of vehicles and maps, and gives the opportunity to edit new maps and vehicles, making it a valuable basis for racing tasks in \acrshort{ad}. Furthermore, the dynamical physics of the vehicles is realistic. Besides that, it lacks complexity, because it does not include pedestrians or intersections.
TORCS is a commonly used driving simulator for \acrshort{rl}~\cite{loiacono_learning_2010,wang_deep_2019}.

\section{Challenges, Recommendations and Outlook}
\label{sec:discussion}
\subsection{Challenges}\label{subsec:challenges}

Generative models share complex and interrelated challenges regarding safety, interpretability, and practical deployment in real-world driving environments.

\paragraph{Safety}
In automotive applications, guidelines such as \acrshort{sotif} \cite{iso_pas_2019} standardize preparations for unforeseen outcomes, incorporating verification and validation akin to rule-based models. 
The opacity of \acrshortpl{dnn} introduces vulnerabilities exploited by adversarial attacks \cite{xie_improving_2019}. Mitigation strategies involve training networks against such adversarial examples \cite{carmon_unlabeled_2019}. 
Incorporating constraints aids in learning knowledge priors \cite{wormann_knowledge_2023}, ensuring predictions remain within defined boundaries. Plausibility-based approaches maintain consistency with physical and environmental constraints, providing parallel safety layers \cite{vivekanandan_plausibility_2022, 
%vivekanandan_ki-pmf_2024
}. 
Beyond use-case based testing, simulation-based methodologies, such as \acrshort{dm}s coupled with \acrshort{rl} frameworks used by GAIA-1 \cite{hu_gaia-1_2023} presented in ~\Cref{subsec:world-models}, enhance safety validation and allow extensive in-simulation testing. Integrating these with established standards bridges traditional safety protocols and the complexities of advanced AI systems. 
Methods like Hamilton-Jacobi (HJ) reachability analysis \cite{chen_hamiltonjacobi_2018} and \acrshortpl{cbf} serve as safety filters or certificates \cite{wabersich_data-driven_2023, hsu_safety_2023} for validating the outcome of learning-based methods, as illustrated by the diffusion-based planners CoBL-Diffusion \cite{mizuta_cobl-diffusion_2024} and SafeDiffuser \cite{xiao_safediffuser_2023} in ~\Cref{sec:planning}. However, computational constraints make their application challenging in unpredictable environments. Recent advancements combine \acrshortpl{cbf} with predictive control \cite{wabersich_predictive_2023} or employ hypernetworks to estimate safe sets in real time based on environmental observations \cite{derajic_learning_2025}.

\paragraph{Interpretability}
Interpreting concepts within generative models is challenging, particularly because these models often contain millions or even billions of parameters.
This opacity poses risks in safety-critical domains, such as \acrshort{ad}, because the system's behavior in response to inputs can be difficult to predict.
Interpretability paradigms are generally divided into four categories: behavioral, attributional, concept-based, and mechanistic \cite{bereska_mechanistic_2024}.
Behavioral interpretability addresses how models respond under various conditions and scenarios.
Attributional interpretability highlights the most influential input features driving the model's output.
This includes methods generating synthetic scenarios coupled with saliency analysis to attribute anomalous detections to specific scenario elements \cite{atakishiyev_explainable_2024}.
While behavioral and attributional interpretability have seen limited application in \acrshort{ad}, larger models pose greater challenges that are more effectively addressed by concept-based and mechanistic interpretability.
Concept-based interpretability relies on semantically meaningful latent factors, while mechanistic interpretability reverse-engineers a model’s internal processes to reveal its decision-making logic.
However, the latter two approaches remain largely underexplored for \acrshort{ad}.
To the best of the authors’ knowledge, only one work \cite{tas_words_2025}, presented in ~\Cref{sec:trajectory_prediction} on Trajectory Forecasting, has applied both concept-based and mechanistic interpretability to \acrshort{ad}.
As models continue to grow in scale, we expect more work to focus on these interpretability paradigms.

\paragraph{Real-time Feasibility}
Efficient operation of generative models within real-time constraints is crucial for autonomous systems. Xu et al.~\cite{xu_-device_2024} review LLM optimization strategies for training and inference, including parameter sharing, modular architectures, quantization, pruning, knowledge distillation, and low-rank factorization \cite{gu_minillm_2024}.
By using Conditional Flow Matching, Ye and Gombolay~\cite{ye_efficient_2024} achieved up to a $100\times$ speedup  for a prediction task, while Gode et al. \cite{gode_flownav_2024} applied the technique for efficient robot navigation. 
Efficiency can also be improved by moving from dense to sparse representations. Tang et al.~\cite{tang_sparseocc_2024} use sparse transformer-based representations to reduce computational overhead by compressing latent variables. Similarly, the \acrshort{e2e} model SparseDrive~\cite{sun_sparsedrive_2024} (cf.~\Cref{sec:end-to-end}) introduces a sparse perception module and parallel motion planner, optimizing efficiency and safety with a hierarchical collision-aware trajectory selection strategy.
Model-based \acrshort{rl} approaches like Hafner et al.~\cite{hafner_dream_2020} and other world models described in ~\Cref{subsec:world-models} reduce online computation requirements by modeling latent dynamics for parallel decision making.
Huang et al. \cite{huang_generative_2025} recently demonstrated generative quantum advantage, proving that quantum processors can efficiently sample distributions that are intractable for classical machines. However, quantum-based approaches fall outside the scope of this survey.
\paragraph{Training and benchmarking}
Beyond safety, interpretability, and real-time feasibility, significant challenges lie in the training and evaluation of generative models. With \acrshort{il}-based training, performance strongly depends on the quality and diversity of the underlying datasets. Large expert-labeled datasets must comprehensively represent real-world scenarios, including rare corner cases, while minimizing bias toward specific distributions. 
Data augmentation, such as introducing artificial shifts and rotations, can partially mitigate covariate shift (cf.~\Cref{par:il}) by training models to recover from suboptimal positions or orientations.
To counter the predominance of straight-driving scenarios, training datasets can be rebalanced to avoid certain situations being overrepresented, encouraging the model to learn more diverse driving behaviors. 
To address these issues in training, cost functions can be integrated alongside \acrshort{il}, such as the \acrshort{e2e} model HydraMDP \cite{li_hydra-mdp_2024} presented in~\Cref{sec:end-to-end}. Moreover, incorporating world models has shown notable performance gains \cite{ma_unleashing_2024}. 
Although \acrshort{rl} methods are becoming increasingly scalable, efficient, and robust across domains \cite{hafner_mastering_2024, espeholt_impala_2018}, they struggle with sparse rewards and weak gradient signals in \acrshort{e2e} training \cite{chen_end--end_2024}. 
\\
The limitations of imitation-based training are closely tied to how models are evaluated: most generative planners rely on open-loop benchmarking, which does not capture the closed-loop feedback essential for realistic driving assessment.
Open-loop evaluation compares a motion planner's output to expert ground truth without considering vehicle or agent effects, whereas closed-loop evaluation includes dynamics and interactions, directly giving feedback on the driving behavior \cite{hallgarten_can_2024}. 
Many \acrshort{sota} generative motion planning models presented in ~\Cref{sec:end-to-end} are optimized for open-loop driving \cite{leonardis_genad_2025, sun_sparsedrive_2024, weng_para-drive_2024, tian_drivevlm_2024}. 
Due to this training paradigm, imitation-based planners lack the robust generalization exhibited by rule-based methods in closed-loop evaluations \cite{dauner_parting_2023}, and often fail to outperform simpler methods under closed-loop conditions \cite{dauner_navsim_2024}. Although \acrshort{e2e} models rarely benchmark against traditional approaches \cite{sun_sparsedrive_2024, hu_planning-oriented_2023, hwang_emma_2024, liao_diffusiondrive_2024}, the research community widely recognizes the need for closed-loop optimization to ensure effective evaluation. However, closed-loop evaluation is limited by a lack of realistic, scalable simulators. 
CARLA \cite{dosovitskiy_carla_2017} offers control but faces sim-to-real limits. Recent simulators introduce closed-loop metrics to open-loop datasets \cite{dauner_navsim_2024}, enhance scenario diversity \cite{yang_drivearena_2024}, and enable photorealistic evaluation \cite{zhou_hugsim_2024}. 

An open gap is the lack of standardized closed-loop benchmarks widely adopted across motion planning and \acrshort{e2e} driving. As a result, generative models are evaluated on heterogeneous datasets, simulators, and metrics with limited scenario diversity, hindering robust comparison of architectures and training strategies. Evaluations often rely on a single dataset or simulator with little attention to sim-to-real gaps and distribution shift under limited scenario diversity, further reducing comparability across studies.
Therefore, trustworthy GenAI in \acrshort{ad} requires standardized, safety-aware closed-loop benchmarks for rigorous evaluation. Although no unified benchmark currently exists, advances in world models and scenario generation are narrowing this gap. %are beginning to narro this gap.}

\begin{figure}
    \centering \includegraphics[width=\columnwidth]{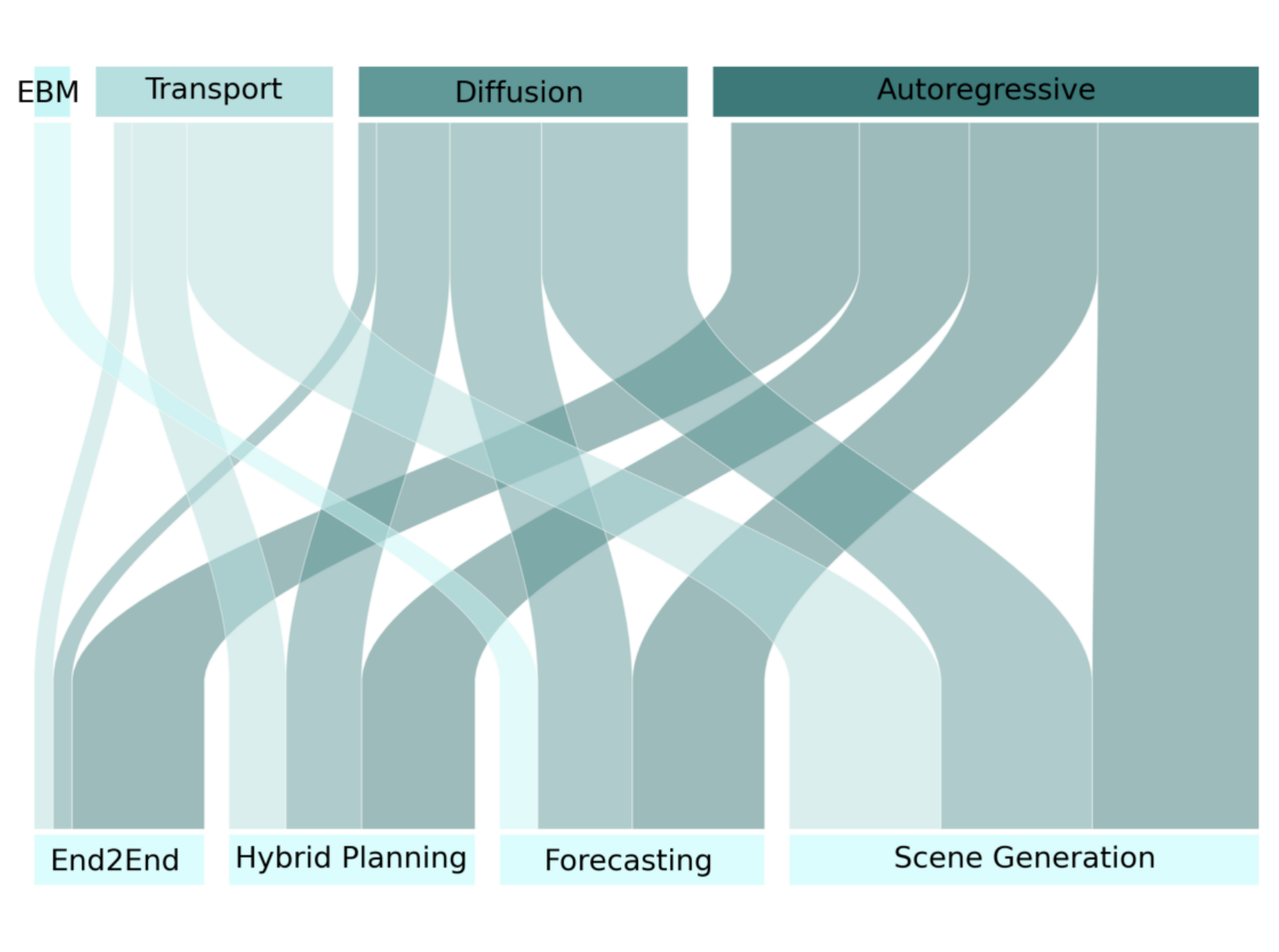}
    \caption{Connection between taxonomy of generative learning approaches in Figure~\ref{fig:taxonomyGenL} and \acrshort{ad} tasks Figure~\ref{fig:ad_stack}. The ribbon width is proportional to the frequency of generative learning approaches contributing to the tasks as noted in Tables \ref{tab:scene-scenario-models}, \ref{tab:trajectory-forecasting-models}, \ref{tab:mot_plan_overview} and \ref{tab:e2e-models}.}
    \label{fig:tax2tasks}   
\end{figure}

\subsection{Advantages and Disadvantages of Generative Paradigms}\label{subsec:advantages}

Various generative models, including \acrshortpl{dm}, transport-based methods like \acrshortpl{vae} and \acrshortpl{gan}, autoregressive Transformers or \acrshortpl{llm}, and \acrshortpl{ebm}, have been explored in the context of \acrshort{ad}, each offering distinct strengths and limitations. 
Figure~\ref{fig:tax2tasks} summarizes their prevalence across the reviewed \acrshort{ad} tasks within our taxonomy in Figure~\ref{fig:taxonomyGenL}. The choice of generative paradigm reflects trade-offs between generation quality, controllability, inference speed and temporal consistency, as shown in Table~\ref{tab:GenAI_Properties}. %Importantly, the effectiveness of generative paradigms depend strongly on the underlying representation, as is especially visible in scene and scenario generation.

Autoregressive models are widely adopted across planning, forecasting and scenario generation. Especially visible in scene and scenario generation, the effectiveness of the generative paradigm depends strongly on the underlying representation. Autoregressive models are well suited to sequential outputs such as road graphs, agents, trajectories, occupancy tokens, and future world states, but can suffer from error accumulation and inefficiency in densely populated scenes.
In planning, \acrshortpl{llm} further enable semantic reasoning, language-based interaction, and improved generalization. However, pretrained \acrshortpl{llm} often lack precise spatio-temporal grounding, inference is typically not real-time, and may produce inconsistent or hallucinatory predictions. The autoregressive Transformer paradigm underlying \acrshortpl{llm} has inspired trajectory forecasting and hybrid planning models, using spatio-temporally grounded and computationally more efficient motion representations or tokens. 
While autoregressive models excel at modeling temporal interactions between agents, this comes at the cost of higher sequential inference latency.

\acrshortpl{dm} are mainly used for forecasting, hybrid planning, and scene or scenario generation, where they provide high-fidelity, diverse, and strongly controllable samples.
Conditioning and inpainting make them useful for generating, completing and editing rasterized or latent \acrshort{bev}, map, occupancy, and video representations. 
This makes them attractive for uncertain and safety-critical situations requiring diverse candidate futures. However, diffusion does not by itself guarantee structural validity, since generated scenes may still require vectorization, topological consistency, or temporal coherence checks. Their high computational cost and slow sampling also limit their adoption in real-time \acrshort{e2e} pipelines.

Transport-based methods are less prevalent, but remain attractive when compact representations and efficient inference are required. They are useful for real-time oriented forecasting and trajectory initialization and policy guidance in hybrid planning. For hybrid motion planning, transport-based models are well-suited for efficient trajectory initialization and policy guidance due to their fast inference.
While latent, retrieval-based, and transport-style models support compact, controllable, and efficient generation, their applicability depends on whether safety-relevant details, interactions, and long-horizon dynamics are adequately depicted.

\acrshortpl{ebm} remain comparatively underexplored in \acrshort{ad} and are mainly applied for trajectory forecasting to model uncertain future trajectories, capture agent interactions, and select or refine representative predictions.
Their use remains limited by slow inference, unstable training, and scalability challenges for long sequences.

\subsection{Recommendations}

While GenAI algorithms are advancing rapidly and often outperform traditional approaches, it remains too early to assess their long-term impact.
The first step when using a GenAI model is to select the appropriate underlying architecture for the task, potentially combining multiple generative approaches in a hybrid framework with classical methods.
As our review of hybrid approaches in~\Cref{subsec:hybrid-methods} suggests, methods integrating elements of traditional planning are likely to remain competitive.
The choice of latent space is another crucial aspect, whether it is based on learned features, as in the early VAE world model \cite{ha_world_2018}, or on more structured abstractions such as BEV agent representations and maps, as used in DriveSceneGen \cite{sun_drivescenegen_2024}, BEVWorld \cite{zhang_bevworld_2024} (cf.~\Cref{sec:scene_generation}), and the \acrshort{e2e} models GenAD \cite{leonardis_genad_2025} and BEVDriver \cite{winter_bevdriver_2025} (cf.~\Cref{sec:end-to-end}).
Abstract `engineered' representations provide clear explainability and facilitate rule-based checks, whereas learned representations typically deliver better performance and are more suitable for resource efficient \acrshort{e2e} approaches \cite{chen_end--end_2024}.
However, they are more susceptible to domain shift.

For decision making, our review underlines that most models use autoregressive LLMs, typically using the same representation or token as in scene generation (cf.~\Cref{sec:scene_generation}). Particularly, models incorporating reasoning capabilities from LLMs present a promising future direction.
In the context of planning, traditional RL methods now compete with action transformers or action state-space models.
GenAI approaches appear to be overtaking pure RL solutions, as reflected, for instance, in the CARLA leaderboard at the time of writing. 

Regarding training methods and data, diversity in training objectives and methods generally improves performance.
Similarly, leveraging multiple diverse datasets is preferable to single-source training, provided the datasets are well curated.

\subsection{Threats to Validity}
We acknowledge potential limitations regarding the scope and completeness of this survey. While we review a range of established generative models, rapidly emerging paradigms, like \acrshort{fm}, may not be exhaustively covered. By focusing on distinctive AD tasks, each subject-specific section has explored the latest developments within its domain to ensure comprehensive coverage of emerging generative models in these focus fields.
A further limitation may be the selection of AD tasks reflected in each subject-specific section, which may deviate from universally agreed classifications. However, this survey aligns with the prevailing conventions from the \acrshort{ad} community, which possesses established definitions for these tasks.
Finally, GenAI is a rapidly evolving field. By the time of publication, new architectures or benchmarks may have emerged post-publication, potentially affecting the currency of specific sections. Consequently, this survey represents a snapshot of the current \acrlong{sota}.

\subsection{Outlook}

GenAI solutions for scene generation appear promising, but the domain gap remains an obstacle when replacing real-world data with synthetic data.
Therefore, methods for measuring the domain gap in driving scenarios must be carefully evaluated, and further development of domain-specific safety-aware metrics is essential.
Additionally, detecting hallucinations and filtering illogical outputs will be another important direction for GenAI quality assurance.     

Despite the advances in GenAI for motion planning, leading algorithms in closed-loop training and testing still exhibit measurable accident rates.
This underscores the need for more robust solutions.
Progress will depend on refining testing methods, identifying corner cases, and re-training GenAI planners.
Additionally, testing in open-loop shadow mode is necessary, alongside novel methods for evaluating proposed but unexecuted actions of GenAI agents.

The deployment of LLM-based autonomous decision making on edge devices poses new challenges, primarily due to the resource demands of reasoning models that offer some degree of explainability.
While knowledge distillation shows promise in downsizing LLMs, in-depth studies in the AD domain are yet to come.
\acrlong{e2e} strategies, spanning from sensor input to action, go in a quite opposite direction, but they necessitate the development of novel testing and diagnostic approaches.

GenAI in AD is an emerging field, and this survey offers only an early glimpse of its potential.
Despite numerous challenges and open research questions, the outlook is promising, and we expect that some of the technical trends outlined here will ultimately mature into robust solutions.

%\section*{Acknowledgment}
%
%Thank you all ;)

\renewcommand{\bibfont}{\small}
\printbibliography

\end{document}